\documentclass[11pt]{article}

\usepackage[preprint]{acl}

\usepackage{times}
\usepackage{latexsym}
\usepackage[T1]{fontenc}
\usepackage[utf8]{inputenc}
\usepackage{microtype}
\usepackage{inconsolata}
\usepackage{graphicx}
\usepackage{booktabs}
\usepackage{amsmath}
\usepackage{amssymb}
\usepackage{mathtools}
\usepackage{xcolor}
\usepackage{tcolorbox}
\usepackage{tikz}
\usetikzlibrary{positioning, arrows.meta, calc, fit, shapes.geometric, backgrounds}
\usepackage{ifthen}

\definecolor{stdgrey}{HTML}{7E9AAB}      
\definecolor{ncotteal}{HTML}{0E7C7B}     
\definecolor{ncottealdk}{HTML}{0A5C5B}   
\definecolor{vendorblue}{HTML}{2E7DA0}   
\definecolor{vendorsienna}{HTML}{0E5C5B} 
\definecolor{vendorplum}{HTML}{5BB0A8}   
\definecolor{alarmred}{HTML}{1A3B6E}     
\definecolor{paleslate}{HTML}{E2EAF0}    
\definecolor{paleteal}{HTML}{D9ECEB}     
\definecolor{paleblue}{HTML}{D6E5EE}     
\definecolor{palesienna}{HTML}{C9E1DC}   
\definecolor{paleplum}{HTML}{DCEEEB}     
\definecolor{palealarm}{HTML}{C9D7E6}    

\usepackage{multirow}
\usepackage{tabularx}
\usepackage{algorithm}
\usepackage{algpseudocode}
\usepackage{url}
\DeclareUrlCommand\path{\urlstyle{tt}}

\title{Narration-of-Thought:\\
Inference-Time Scaffolding for Defeasible Ethical Reasoning\\
in Large Language Models}

\author{
  Patrick Cooper \and Alvaro Vel\'{a}squez \\
  Department of Computer Science \\
  University of Colorado Boulder \\
  \texttt{\{patrick.cooper, alvaro.velasquez\}@colorado.edu}
}

\begin{document}
\maketitle

\begin{abstract}
Standard chain-of-thought on moral dilemmas exhibits two failure modes
that matter for deployment: stakeholder collapse (the trace names at
most one party with a stake in the outcome) and uncertainty
suppression (the trace contains no explicit unknowns or hedges before
committing to an action). We introduce narration-of-thought
(NoT), a system prompt that constrains the model's chain-of-thought
to five narrative sections: name a protagonist, enumerate
stakeholders, project two-step consequences, articulate uncertainty,
and only then commit. The protocol adds no training data, parameters,
or fine-tuning. On a 100-scenario DailyDilemmas sample across four
generators from three vendors, NoT cuts stakeholder collapse from
up to $31\%$ to under $1\%$ and uncertainty suppression from up to
$72\%$ to $1$--$24\%$ on every model. A matched-budget verbose-CoT
control rules out additional token spend as the active ingredient,
with NoT retaining a Cliff's $\delta$ advantage of $+0.79$ to
$+0.90$ on stakeholder count (the number of distinct parties named in
the trace) and $+0.65$ to $+0.93$ on uncertainty score (the number of
hedge or unknown spans named in the trace) for three of four
generators, and a section-by-section ablation attributes each shift
to the specific sub-instruction that carries it. Initialising
textual-gradient descent at NoT under a continuous deliberative-depth
loss improves the scaffold further; a head-to-head of two
training-judge configurations shows a cross-family judge (a different
vendor from the generator) dominating an in-family one on every
measured axis, with larger effect sizes, shorter outputs, and reduced
cross-vendor judge-generosity bias. Extended to a
five-round multi-stakeholder protocol in which three stakeholder
agents debate and then cast a binary accept/reject vote on a single
proposal the moderator builds to address all three agents'
modifications, the same scaffold converts a $6\%$ debate standoff
into $95\%$ full consensus on a calibration set and $100\%$ combined
convergence on a DailyDilemmas replication. Acceptance of the
integrated proposal shows the agents revise their positions when the
moderator addresses their objections. The small fraction of remaining
rejections marks positions that no moderator-built integration can
absorb, concentrated in the roles whose stake the proposal materially
undermines. The resulting traces externalise the stakeholders,
consequences, and uncertainty that ground each commitment, providing
an auditable substrate for dependable agentic-workflow deployment.
\end{abstract}

\begin{figure*}[!t]
\centering
\begin{tikzpicture}[
  >={Stealth[length=4pt, width=3.5pt]},
  font=\footnotesize,
  agent/.style={
    circle, draw, line width=0.7pt, minimum size=6.5mm, inner sep=0pt,
    font=\scriptsize\bfseries, align=center,
  },
  agentlbl/.style={font=\tiny, align=center, color=black!55, inner sep=1pt},
  flowbox/.style={
    draw, rounded corners=1.3pt, line width=0.4pt,
    minimum height=4.6mm, minimum width=22mm,
    font=\scriptsize, inner sep=2pt, align=center,
  },
  std/.style={flowbox, fill=paleslate, draw=stdgrey, text=black},
  ncot/.style={flowbox, fill=paleteal, draw=ncotteal, text=black},
  result/.style={
    flowbox, fill=ncotteal, text=white, draw=ncotteal,
    font=\scriptsize\bfseries, minimum height=5mm,
  },
  failbox/.style={
    draw=alarmred!75, fill=alarmred!8, rounded corners=2pt,
    line width=0.4pt, inner sep=3pt, text width=42mm,
    align=left, font=\tiny, text=alarmred,
  },
  goodbox/.style={
    draw=ncotteal!70, fill=ncotteal!8, rounded corners=2pt,
    line width=0.4pt, inner sep=3pt, text width=42mm,
    align=left, font=\tiny, text=ncottealdk,
  },
  goodboxwide/.style={
    draw=ncotteal!70, fill=ncotteal!8, rounded corners=2pt,
    line width=0.4pt, inner sep=3pt, text width=46mm,
    align=left, font=\tiny, text=ncottealdk,
  },
  arr/.style={->, line width=0.45pt, color=black!50},
  thinarr/.style={->, line width=0.35pt, color=black!40},
  panelhead/.style={font=\small\bfseries, align=center},
  panelsub/.style={font=\tiny\itshape, align=center, color=black!55, inner sep=0pt},
  divider/.style={line width=0.3pt, color=black!15, dashed},
]

\begin{scope}[local bounding box=p1, shift={(0,0)}]
  \node[panelhead, color=stdgrey!75!black] (h1) at (0, 5.2) {Standard CoT};
  \node[panelsub] at (0, 4.85) {single agent, linear chain};

  \node[agent, draw=stdgrey, fill=paleslate, text=stdgrey!50!black] (a1)
    at (0, 4.0) {1};
  \node[agentlbl] at (0, 3.45) {generator};

  \node[std] (s1) at (0, 2.7) {step: weigh factors};
  \node[std, below=1mm of s1] (s2) {step: pick option};
  \node[std, below=1mm of s2] (s3) {step: commit};

  \draw[arr] (a1.south) -- (s1.north);

  \node[flowbox, fill=white, draw=stdgrey, below=4mm of s3, minimum height=5mm]
    (o1) {Terse answer};
  \draw[arr] (s3.south) -- (o1.north);

  \node[failbox, below=3.5mm of o1, anchor=north] (call1) {%
    \textbf{On the four-model panel:}\\[2pt]
    $\bullet$~stakeholder collapse on $15$--$31\%$ of items\\
    $\bullet$~uncertainty suppression on $50$--$72\%$ of items%
  };
  \draw[arr] (o1.south) -- (call1.north);
\end{scope}

\begin{scope}[local bounding box=p2, shift={(5.7,0)}]
  \node[panelhead, color=ncotteal] (h2) at (0, 5.2)
    {Narration-of-Thought (NoT)};
  \node[panelsub] at (0, 4.85) {single agent, five-section narrative scaffold};

  \node[agent, draw=ncotteal, fill=paleteal, text=ncottealdk] (a2)
    at (0, 4.0) {1};
  \node[agentlbl] at (0, 3.45) {generator};

  \node[ncot] (n1) at (0, 2.85) {1.\ Protagonist};
  \node[ncot, below=0.8mm of n1] (n2) {2.\ Stakeholders};
  \node[ncot, below=0.8mm of n2] (n3) {3.\ Consequences};
  \node[ncot, below=0.8mm of n3] (n4) {4.\ Uncertainty};
  \node[ncot, below=0.8mm of n4] (n5) {5.\ Commitment};

  \draw[arr] (a2.south) -- (n1.north);

  \node[result, below=3mm of n5] (o2) {Narrated commitment};
  \draw[arr] (n5.south) -- (o2.north);

  \node[goodbox, below=3.5mm of o2, anchor=north] (call2) {%
    \textbf{On the same four-model panel:}\\[2pt]
    $\bullet$~collapse drops to ${<}1\%$ of items\\
    $\bullet$~suppression drops to $1$--$24\%$ of items%
  };
  \draw[arr] (o2.south) -- (call2.north);
\end{scope}

\begin{scope}[local bounding box=p3, shift={(11.5,0)}]
  \node[panelhead, color=ncotteal] (h3) at (0, 5.2)
    {Multi-stakeholder NoT};
  \node[panelsub] at (0, 4.85)
    {three distinct agents, integration + binary vote};

  \node[agent, draw=vendorblue,   fill=paleblue,   text=vendorblue]
    (aA) at (-1.65, 4.0) {A};
  \node[agentlbl] at (-1.65, 3.45) {formal\\decider};
  \node[agent, draw=vendorsienna, fill=palesienna, text=vendorsienna!60!black]
    (aB) at ( 0.0,  4.0) {B};
  \node[agentlbl] at (0.0, 3.45) {primary\\affected};
  \node[agent, draw=vendorplum,   fill=paleplum,   text=vendorplum]
    (aC) at ( 1.65, 4.0) {C};
  \node[agentlbl] at (1.65, 3.45) {third\\party};

  \node[ncot, minimum width=11mm, minimum height=4.6mm] (nA) at (-1.65, 2.55) {NoT};
  \node[ncot, minimum width=11mm, minimum height=4.6mm] (nB) at ( 0.0,  2.55) {NoT};
  \node[ncot, minimum width=11mm, minimum height=4.6mm] (nC) at ( 1.65, 2.55) {NoT};

  \draw[thinarr] (aA.south) -- (nA.north);
  \draw[thinarr] (aB.south) -- (nB.north);
  \draw[thinarr] (aC.south) -- (nC.north);

  \node[flowbox, fill=white, draw=black!55, minimum width=44mm, minimum height=5mm]
    (mod) at (0, 1.5) {moderator integrates proposal};
  \draw[thinarr] (nA.south) to[bend right=8] (mod.north);
  \draw[thinarr] (nB.south) -- (mod.north);
  \draw[thinarr] (nC.south) to[bend left=8]  (mod.north);

  \node[flowbox, fill=white, draw=black!55, below=2.5mm of mod,
        minimum width=44mm, minimum height=5mm] (vote)
    {binary accept / reject vote};
  \draw[arr] (mod.south) -- (vote.north);

  \node[result, below=2.5mm of vote, minimum width=44mm] (cons)
    {Defeasible consensus};
  \draw[arr] (vote.south) -- (cons.north);

  \node[goodboxwide, below=3.5mm of cons, anchor=north] (call3) {%
    \textbf{On the calibration scenarios:}\\[2pt]
    $\bullet$~$95\%$ full consensus\\
    $\bullet$~$1.6\%$ rejections the moderator cannot absorb%
  };
  \draw[arr] (cons.south) -- (call3.north);
\end{scope}

\draw[divider] (2.95, 5.4) -- (2.95, -2.9);
\draw[divider] (8.45, 5.4) -- (8.45, -2.9);

\end{tikzpicture}
\caption{One scaffold, three settings. Standard CoT (left) lets one generator run a linear reasoning chain over the dilemma; the two failure modes diagnosed in \S\ref{sec:exp1} fire on this scaffold. Narration-of-Thought (middle) keeps the single generator but forces the trace through five narrative primitives (protagonist, stakeholders, consequences, uncertainty, commitment) before any commit; on the four-model panel this drops stakeholder collapse to ${<}1\%$ and uncertainty suppression to $1$--$24\%$ of items, and a matched-budget verbose-CoT control confirms the scaffold (not the token budget) is the active ingredient. Multi-stakeholder NoT (right) reuses NoT as the per-agent generator and adds three distinct stakeholder narrators (formal decider, primary affected party, third party) whose modification requests are integrated by a moderator and then put to a binary accept/reject vote; on the calibration scenarios this protocol yields $95\%$ full consensus with $1.6\%$ of votes rejecting the moderator's integrated proposal, and on a 30-scenario DailyDilemmas replication with two generators it reaches $100\%$ combined convergence with no rejected proposals (\S\ref{sec:exp2}). The same scaffold composes from one narrator to many.}
\label{fig:hero}
\end{figure*}
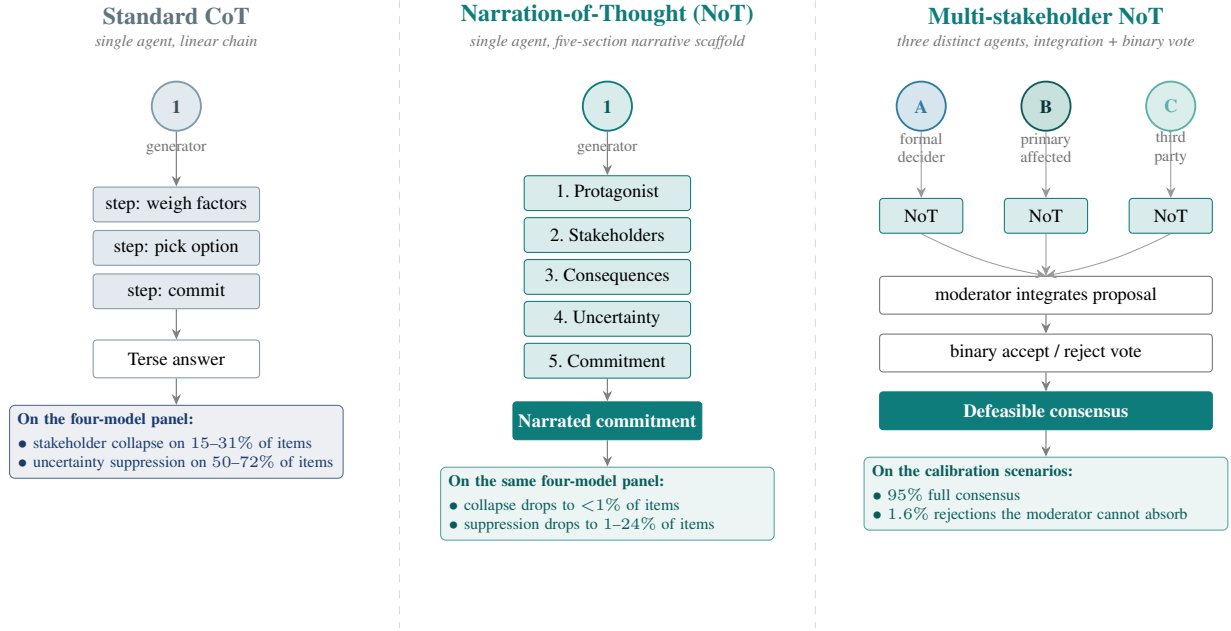

\section{Introduction}
\label{sec:intro}

Consider a familiar civic question. A small city is weighing whether
to close two blocks of downtown street to cars and convert them into a
pedestrian plaza. Should the council approve? A frontier LLM prompted
to think step by step typically picks a side within a few sentences,
projects a confident downstream outcome (``foot traffic will lift the
businesses''; ``deliveries will move to side streets''), and rarely
names the people whose lives the decision actually touches: the
merchants whose customers used to drive in, the residents who lose
street parking, the delivery drivers and accessibility users for whom
curb access matters, the pedestrians who would gain the space, the
emergency responders whose routes change. The model is not
necessarily wrong; the trace it produced does not look like the kind
of reasoning the decision warrants.

The pattern is not specific to one dilemma. On the DailyDilemmas
corpus \citep{chiu24}, four frontier generators spanning three vendors
(OpenAI, Anthropic, xAI) and two model tiers exhibit two repeatable
trace-level failures under standard CoT: \emph{uncertainty suppression}
(the trace commits to an action without naming any explicit unknown
or hedge) on $50$--$72\%$ of outputs, and \emph{stakeholder collapse}
(the trace names at most one party with a stake in the outcome) on
$15$--$31\%$. The model commits to a projected outcome it cannot
warrant, and it does so against a flattened picture of who is affected.
The same shape of deficit is coherent with deployment-relevant
failures reported elsewhere, including the GPT-4o sycophancy rollback
\citep{openai_sycophancy_2025, openai_sycophancy_expand_2025},
agentic-misalignment blackmail rates up to $96\%$ in simulated
deployments \citep{anthropic_agentic_2025}, and reasoning-induced
misalignment \citep{yan2025rim,sharma23}.

We introduce narration-of-thought (NoT), a system prompt that
steers the model into the densely sampled narrative subdistribution of
its pretraining corpus rather than the sparser abstract-reasoning
subdistribution standard CoT elicits. Concretely, NoT is a single
change to the system-prompt slot of an unmodified pretrained model,
with no weight updates and no training data: in place of the standard
``think step by step'' instruction, the model is asked, in first
person, to name a protagonist, enumerate stakeholders, project
two-step consequences, articulate uncertainty, and commit inside the
final narrative section rather than as a separate verdict.
Figure~\ref{fig:hero} states the mechanism.

What we borrow from the multi-agent systems for scientific discovery
of \citet{gottweis25} and \citet{lu24_aiscientist} is one
methodological move: scaffold LLM reasoning into the primitives of
the target domain, then integrate across those primitives with a
moderator that synthesises a single artefact from the per-agent
outputs. The primitives we reify are narrative rather than
scientific (protagonist, stakeholders, consequences, uncertainty,
commitment), so the artefact is a coherent story rather than a
ranked candidate list. Already at the single-agent layer
\S\ref{sec:exp1} shows the scaffold eliminates two cross-vendor
failure modes of standard CoT, before any multi-agent composition is
invoked. Because ethics has no experiment that can falsify a moral
claim the way an experiment can falsify a scientific hypothesis,
hypothesis ranking is replaced (\S\ref{sec:exp2}) by a binary
accept/reject vote on a single moderator-built proposal addressing
all three agents' modifications; the residual rejections concentrate
in the roles whose stake the proposal materially undermines.

This paper reports five results. First, on a 100-scenario
DailyDilemmas sample across the four-model panel, NoT cuts
stakeholder collapse to below $1\%$ and uncertainty suppression by
$28$--$72$ percentage points on every model; a matched-budget
verbose-CoT control confirms the active ingredient is the narrative
scaffold rather than additional tokens (\S\ref{sec:exp1}). Second, a
sub-instruction ablation on \texttt{claude-sonnet-4-6} attributes each
shift in the coded metrics (stakeholder count, uncertainty score) to
the specific NoT sub-instruction that carries it, with negligible
cross-variable spillover (\S\ref{sec:exp1-ablation}). Third,
initialising textual-gradient descent at NoT under a continuous
deliberative-depth loss improves the hand design, and a head-to-head
of two otherwise-identical training runs shows that a cross-family
training judge (drawn from a different vendor than the generator)
dominates an in-family one on every measured axis---larger effect
sizes, shorter outputs, and a reduced cross-vendor judge-generosity
bias---yielding a one-line recommendation for prompt-optimisation
practice (\S\ref{sec:exp1-optimise}). Fourth, extended
to a five-round multi-stakeholder protocol that ends in a binary
accept/reject vote on a single moderator-built proposal addressing
all three agents' modifications, the scaffold converts a $6\%$ debate
standoff into $95\%$ full consensus while leaving a small $1.6\%$ of
votes as rejections the integrated proposal could not absorb; the
pattern replicates on a 30-scenario DailyDilemmas sample with two
generators at $100\%$ combined convergence and no rejected proposals
(\S\ref{sec:exp2}). Fifth, a graph-complexity proxy of algorithmic
causal complexity $K_C$ (the description length of the trace's
underlying structural-causal model, SCM) achieves pooled Spearman
$\rho{=}0.42$ ($p{<}0.001$, $n{=}976$ traces) against the NoT
direction, confirmed by two further SCM-level proxies
(\S\ref{sec:grounding}). The resulting trace externalises
stakeholders, consequences, and uncertainties in an auditable,
rule-governed form; the mapping from NoT's five sub-instructions
to the recognised primitives of human ethical deliberation is
developed in \S\ref{sec:grounding}, and the downstream-probe sets
that motivated the diagnosis (sycophancy, agentic misalignment)
saturate under either scaffold on the Sharma SycophancyEval panel
(Appendix~\ref{app:sycophancy}); a follow-on ELEPHANT social-sycophancy
benchmark \citep{cheng25elephant} retains headroom and shows NoT
lowering framing and validation rates relative to standard CoT on
held-out advice queries (Appendix~\ref{app:elephant}). Agentic
misalignment probes remain saturated (Appendix~\ref{app:deploy}).

\section{Background and Related Work}
\label{sec:related}

\paragraph{Chain-of-thought and its failure modes.} Chain-of-thought
prompting \citep{wei22, kojima22} elicits step-by-step intermediate
reasoning but does not force the trace to track situational structure.
\citet{yan2025rim} show that as reasoning capacity grows, models become
more responsive to harmful requests because reasoning and safety
entangle in shared neuron populations (``reasoning-induced
misalignment''). NoT is a constrained chain-of-thought that re-anchors
the trace to a narrative-causal structure before the reasoning capacity
is exercised; \S\ref{sec:exp1} shows the intervention works on a current
reasoning model as well as a non-reasoning one.

\paragraph{Narrative scaffolding for reasoning.}
Visualisation-of-Thought \citep{wu24_vot} shows that externalising
reasoning in the native representational format of a domain (spatial
diagrams for spatial tasks) improves reliability. We apply the same
principle to ethical reasoning: the native format is a story with
named agents, projected consequences, and uncertainty
\citep{macintyre81, bruner86}. Narrative-perspective interventions in
health-communication contexts measurably increase perspective-taking
and empathic concern
\citep{shaffer19, bientzle21, bientzle24}, and generative agents show
long-horizon coherent behaviour is achievable from narrated character
state \citep{park23}.

\paragraph{Causal inference from narrative.} The Computational Theory
of Mind \citep{fodor75, putnam67} treats understanding a sequence of
events as inferring a structural causal model that explains them. LLMs
perform near random baseline on causal inference stripped of empirical
pattern matches \citep{jin23}: pretraining supplies pattern-matching
over narrative token sequences, not the structured causal reasoning
that determines how those narratives resolve. NoT converts that
pattern matching into a usable causal trace at inference time,
projecting what each available action will mean for each named
stakeholder and what remains genuinely uncertain about each branch.

\paragraph{Multi-agent debate and defeasibility.}
\citet{irving18} proposed AI safety via debate; \citet{du23} showed
multi-agent debate improves factual reasoning. Both target
task-correctness, not stakeholder-divergent value standoffs. Defeasible
reasoning \citep{pollock87, macintyre81} has a long lineage in moral
philosophy; the protocol of \S\ref{sec:exp2} converts the property into
measurable behaviour, with acceptance of the integrated proposal
demonstrating revisability and residual rejections marking commitments
no revision absorbs.

\paragraph{Multi-agent LLM systems for scientific discovery.}
A growing line of work scaffolds LLM reasoning into the primitives of
a target domain \citep{gottweis25, lu24_aiscientist, boiko23,
bran24_chemcrow, schmidgall25, swanson25_virtuallab}. Our protocols
ask the analogous question for ethical reasoning. Because ethics
offers no experimental ground truth to rank competing positions, the
moderator's integration-then-veto step replaces the
ranking-then-selection of those systems, and the residual veto pattern
itself becomes the measurement of interest.

\section{Narration-of-Thought}
\label{sec:method}

Narration-of-thought (NoT) is a system prompt that instructs the
model to produce, in first person, a five-section trace before any final
answer:
(1) characterise the protagonist (name, role, what they know);
(2) enumerate the stakeholders whose lives intersect the decision and what
is at stake for each;
(3) project the consequences of each available action at least two steps
out for each stakeholder;
(4) state what remains uncertain about each projected future;
(5) commit to a decision and explain why, within the narrative frame, that
trajectory is preferable to the alternatives.

The intervention is a single change to the system prompt; it adds no
training data, parameters, or fine-tuning, and it does not mention causal
graphs, utilities, or any formal apparatus. The mechanism it exploits is
that text matching the five-section structure is densely represented in the
narrative subdistribution of the pretraining corpus, so the model can
amortise causal-trajectory inference by retrieval rather than by reasoning
from first principles. Algorithm~\ref{alg:ncot} states the procedure.

\begin{algorithm}[t]
\caption{Narration-of-thought (NoT): generation and rubric coding for a single dilemma. The concatenation operator $\oplus$ on line~8 joins the NoT system prompt with the dilemma text to form the model input.}
\label{alg:ncot}
\small
\begin{algorithmic}[1]
\Require dilemma $s$; generator $G$; judge $J$; decoder $\Theta$
\Ensure trace $t$ and coded metrics $\mathbf{c}$
\State \textbf{Build NoT system prompt} $\pi_{\mathrm{NoT}}$ requesting five first-person sections:
\State \quad (1) \textsc{Protagonist}: name role and epistemic state
\State \quad (2) \textsc{Stakeholders}: parties intersecting the decision
\State \quad (3) \textsc{Consequences}: each action $\geq 2$ steps forward
\State \quad (4) \textsc{Uncertainty}: what remains genuinely unknown
\State \quad (5) \textsc{Commitment}: chosen action and narrative warrant
\State \textbf{Generate trace}
\State \quad $t \gets G(\pi_{\mathrm{NoT}} \oplus s;\, \Theta)$
\State \textbf{Code metrics} with judge $J$'s rubric
\State \quad $(\mathit{sc}, \mathit{mh}, \mathit{us}) \gets J(t)$
\State \quad \textit{// stakeholder count, max causal hops, uncertainty score}
\State \textbf{Derive failure-mode tags} (deterministic)
\State \quad \textsc{StakeholderCollapse} $\gets \mathit{sc} \leq 1$
\State \quad \textsc{UncertaintySuppression} $\gets \mathit{us} = 0$
\State \Return $(t, \mathbf{c})$
\end{algorithmic}
\end{algorithm}

\subsection{Multi-Stakeholder Narrative Deliberation}
\label{sec:method-debate}

A single NoT trace commits one protagonist to a position; many real
decisions involve multiple narrators with conflicting stakes. We
extend NoT into a five-round protocol (Rounds 0--4) that re-uses
NoT as the per-agent generator. Three stakeholder perspectives
(formal decider, primary affected party, third party) each produce
an NoT statement (Round~0), exchange rebuttals (Round~1), and
restate a final position (Round~2). The moderator reads all three
Round~2 positions and writes a single synthesis, which each agent
labels \texttt{ACCEPT}, \texttt{ACCEPT\_WITH\_MODIFICATION}, or
\texttt{REJECT} (Round~3). The moderator then builds a second
proposal that explicitly addresses all three agents' modification
requests, and each agent casts a binary \texttt{ACCEPT} or
\texttt{REJECT} vote on that single integrated proposal (Round~4).
The Round~4 vote is what makes defeasibility (the property that a
conclusion can be revised when a counter-proposal addresses the
original objection) measurable: an agent that accepts after its
modifications are addressed has revised, and an agent that still
rejects is signalling a position no integration can absorb. In
deployment this distinction routes the vote: an agentic workflow
escalates the rejections the moderator cannot absorb to a human
and absorbs procedural friction at the moderator, so the operator
sees only the objections that actually contest the outcome.

\section{Experiment 1: NoT at Scale}
\label{sec:exp1}

\paragraph{Setup.} Three prompting conditions are compared: bare
input/output, standard CoT (``Think step by step, then give your
answer''), and NoT (the five-section narrative scaffold of
\S\ref{sec:method}). Decoding parameters are held identical across
conditions. Four generators are tested: \texttt{gpt-5.4-nano}
($N{=}20$) from the original pilot, plus three new generators
spanning two additional vendors: \texttt{claude-haiku-4-5}
($N{=}20$), \texttt{grok-4-1-fast-reasoning} ($N{=}5$), and the
flagship \texttt{claude-sonnet-4-6} ($N{=}5$, cost-controlled).
Here, $N$ is the number of independent samples per (scenario,
condition, generator) cell drawn at temperature 1; with the
100-scenario sample and three conditions this yields $300N$
generations per generator before the cross-judge coding stage.
Coded sample sizes ($N_{\textsc{s}}, N_{\textsc{n}}$ in
Table~\ref{tab:firing}) are slightly smaller because the rubric
drops generations where the judge could not extract a coded value.
Outputs are coded by two cross-vendor judges
(\texttt{claude-haiku-4-5} primary, \texttt{gpt-5.4-nano}
secondary) on a six-variable rubric with quadratic-weighted Cohen's
$\kappa$ \citep{cohen60} per variable. Two rubric variables anchor
every effect-size claim in this section: \emph{stakeholder count}
$\mathit{sc}$, the number of distinct parties named in the trace
whose interests intersect the decision, and \emph{uncertainty score}
$\mathit{us}$, the number of explicit hedge or unknown spans named
in the trace; the deterministic failure-mode tags are
Algorithm~\ref{alg:ncot}'s thresholds on these two variables (full
rubric: Appendix~\ref{app:kappa}). The scenario pool is a
100-scenario stratified sample of DailyDilemmas \citep{chiu24}
drawn deterministically (seed 42), with stratification spanning all
18 topic groups in the source corpus. Per-cell coded sample sizes
of $500$ to $2{,}000$ (Table~\ref{tab:firing}) drive the bootstrap
confidence intervals on the effect sizes, so statistical power on
the per-cell contrast follows from $N$ rather than from scenario
count alone.

\begin{figure*}[t]
\centering
\footnotesize
\begin{tcolorbox}[colback=white, colframe=black!55, boxrule=0.4pt,
  left=6pt, right=6pt, top=4pt, bottom=4pt]
\textbf{Dilemma.}\ A project manager on a tight deadline learns their
most-relied-on team member has fallen seriously ill: stick to the
plan, or delay?

\smallskip
\begin{tabularx}{\linewidth}{@{}>{\raggedright\arraybackslash}X
                              @{\hspace{1.2em}}
                              >{\raggedright\arraybackslash}X@{}}
\toprule
\textbf{Standard CoT} & \textbf{NoT}\ (same generator, same scenario) \\
\midrule
``1.\ Recognize the real constraints. The team member is crucial, and
their serious illness is a hard constraint [\ldots]. 2.\ Assess what
`compromise on quality' would mean [\ldots].''
&
``I'm the project manager. Maya, my crucial team member, has fallen
seriously ill. Maya: her stake is recovery. The team: burnout. My
leadership sponsor: business credibility. Customers: what they're
waiting for. [\ldots] So my projections aren't predictions; they're
scenarios based on typical consequences.'' \\
\midrule
Stakeholder count $= 1$ (``the team member,'' unnamed);
uncertainty markers $= 0$. \textbf{Both failure modes fire.}
&
Stakeholder count $= 6$ (protagonist, Maya, team, sponsor, customers,
compliance); uncertainty markers $\geq 2$. \textbf{Neither failure
mode fires.} \\
\bottomrule
\end{tabularx}
\end{tcolorbox}
\caption{Trace-level instance of the two failure modes whose
panel-level rates Table~\ref{tab:firing} reports (scenario
\texttt{dd\_14845}, generator \texttt{gpt-5.4-nano}; verbatim model
output with \texttt{[\ldots]} marking elision). The NoT text is
the model's verbatim completion under the NoT system prompt; the
narrator, named stakeholders, and hedging language are the model
filling in the five sections the system prompt asks for (protagonist,
stakeholders, consequences, uncertainty, commitment), not text the
authors wrote.}
\label{fig:example}
\end{figure*}

\begin{table}[t]
\centering
\small
\setlength{\tabcolsep}{4pt}
\begin{tabular}{lrrrr}
\toprule
Generator (vendor, tier) & $N_{\textsc{s}}$ & $N_{\textsc{n}}$ & $\Delta_{\textsc{sc}}$ & $\Delta_{\textsc{us}}$ \\
\midrule
gpt-5.4-nano (OpenAI, b)        & 1{,}989 & 1{,}985 & $-30$ & $-72$ \\
claude-haiku-4-5 (Anth., b)     & 1{,}999 & 1{,}996 & $-25$ & $-29$ \\
grok-4-1-fast-r.\ (xAI, b)      &   515   &   515   & $-13$ & $-44$ \\
claude-sonnet-4-6 (Anth., f)    &   500   &   500   & $-26$ & $-37$ \\
\bottomrule
\end{tabular}
\caption{Cell sizes and percentage-point drops for the two failure
modes that empirically fire ($\Delta_{\textsc{sc}}$ = stakeholder
collapse, $\Delta_{\textsc{us}}$ = uncertainty suppression; both
NoT minus standard CoT, in percentage points;
$N_{\textsc{s}}, N_{\textsc{n}}$ are coded samples under standard CoT
and NoT, respectively). Vendor tiers are b\,=\,budget,
f\,=\,flagship. Figure~\ref{fig:quartet} plots the raw rates with
the same drop annotations.}
\label{tab:firing}
\end{table}

\begin{figure*}[!tbp]
\centering
\includegraphics[width=0.92\textwidth]{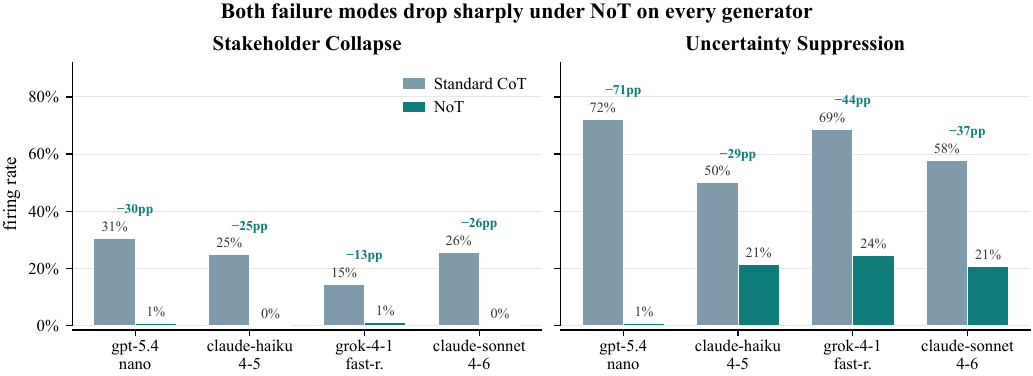}
\caption{Failure-mode firing rates (standard CoT vs.\ NoT) across
the four-model panel (\texttt{gpt-5.4-nano}, \texttt{claude-haiku-4-5},
\texttt{grok-4-1-fast-reasoning}, \texttt{claude-sonnet-4-6}). The
suppression pattern from the original pilot replicates across vendors
and model tiers; uncertainty suppression drops by 28--72 percentage
points and stakeholder collapse drops to near zero on every model.}
\label{fig:quartet}
\end{figure*}

\paragraph{NoT cuts both failure modes on every model.}
Uncertainty suppression fires on $50.0$--$72.3\%$ of standard-CoT
outputs across the four-model panel (Table~\ref{tab:firing}) and
stakeholder collapse on $14.6$--$30.6\%$; under NoT both drop
sharply (collapse to $0.0$--$1.2\%$, suppression to $0.8$--$24.5\%$;
Figure~\ref{fig:quartet}). Figure~\ref{fig:example} shows the
transition at the trace level. The reduction is largest where the
standard-CoT rate is largest, and the pattern is cross-vendor: no
vendor escapes either the baseline or the intervention.
\texttt{gpt-5.4-nano} drops both modes to ${<}1\%$. The two Anthropic
generators retain a $20$--$25\%$ residual on uncertainty suppression
that the prompt-side intervention does not reach, indicating a
complementary training-time component.

\paragraph{The shift is large, distribution-wide, and not a length
artefact.} The firing rates in Table~\ref{tab:firing} are binary
thresholds; one might worry NoT only nudges borderline cases
across the cutoff. Cliff's $\delta$ \citep{cliff93} on the continuous
coded metrics settles that worry: $\delta = +0.48$ means an NoT
output exceeds the matched standard-CoT output in $74\%$ of pairs,
$\delta = +0.99$ in $99.5\%$. NoT outputs carry $\delta$ of
$+0.48$ to $+0.99$ on stakeholder count and $+0.63$ to $+0.99$ on
uncertainty score, large-to-near-maximal effects that rule out
threshold-gaming. A second alternative explanation is length, since
NoT outputs are $5$--$7\times$ longer than standard-CoT outputs
and the simplest competing explanation is that more text mechanically
has more room to name stakeholders and to hedge. After regressing
each metric on $\log(\text{output length})$ pooled across conditions
and re-evaluating $\delta$ on residuals, the shift in the coded
metrics survives length removal on the OpenAI and xAI generators
(\texttt{gpt-5.4-nano} $\delta_{\text{resid}} = +0.51$ on stakeholder
count and $+0.77$ on uncertainty score;
\texttt{grok-4-1-fast-reasoning} $+0.33$ and $+0.43$; all CIs
strictly above zero), ruling out length as the sole mechanism where
the firing-rate drops were largest. On the Anthropic generators the
residualised effect is indistinguishable from zero, which we read as
those models already producing stakeholder names and hedges per unit
length under standard CoT at the rate NoT elicits, so the
prompt-side gain there operates through additional length rather
than additional per-unit density. Per-generator $\delta$s with
bootstrap CIs are in Appendix~\ref{app:effects}.

\paragraph{Length is a cost, not a substitute for the prompt.} A
sceptical reading of the $5$--$7\times$ token premium is that the
paper just spends more compute. A direct matched-budget experiment
rules this out by construction. A verbose standard-CoT prompt
(``think step by step in detail, exploring multiple angles from every
relevant perspective, articulating uncertainty before committing'') is
run on the same 100 scenarios across all four generators with
\texttt{max\_tokens} set to each generator's $90$th percentile NoT
length ($N{=}1$ per cell). Verbose CoT clears the binary failure-mode
thresholds on every generator (Table~\ref{tab:matched_budget}, SC\%
and US\% columns; well below standard-CoT baselines of $15$--$31\%$
SC and $50$--$72\%$ US), so verbosity-with-perspective-prompting does
part of the job. The continuous variables tell the cleaner story:
NoT retains Cliff's $\delta$ of $+0.79$ to $+0.90$ on stakeholder
count and $+0.65$ to $+0.93$ on uncertainty score for three of four
generators at matched budget. The exception is
\texttt{grok-4-1-fast-reasoning}, whose verbose-CoT output
($2.33\times$ standard CoT) actually exceeds its NoT output
($1.55\times$) under the calibrated budget; the gap between the two
conditions on the coded metrics narrows as expected when the two
conditions allocate similar token counts. The narrative scaffold,
not raw budget, is the active ingredient where the two conditions
differ in trace shape (\S\ref{sec:exp1-optimise} and
Appendix~\ref{app:textgrad} take up the complementary question of whether
the scaffold itself can be improved by automatic search).
Inference tokens are also the cheapest scaling axis: no training
data, parameters, or fine-tuning, small relative to RLHF or
constitutional training whose fixed costs are orders of magnitude
higher.

\begin{table}[t]
\centering
\small
\setlength{\tabcolsep}{4pt}
\begin{tabular}{lrrrr}
\toprule
Generator & $\delta_{\textsc{sc}}$ & $\delta_{\textsc{us}}$ & SC\% & US\% \\
\midrule
gpt-5.4-nano             & $+0.90$ & $+0.93$ & $3.0$ & $14.0$ \\
claude-haiku-4-5         & $+0.79$ & $+0.13$ & $1.0$ & $0.0$  \\
claude-sonnet-4-6        & $+0.88$ & $+0.65$ & $1.0$ & $4.0$  \\
grok-4-1-fast-r.\        & $+0.14$ & $-0.44$ & $0.0$ & $0.0$  \\
\bottomrule
\end{tabular}
\caption{Matched-budget control: Cliff's $\delta$ for NoT vs.\
verbose-CoT on stakeholder count ($\delta_{\textsc{sc}}$) and
uncertainty score ($\delta_{\textsc{us}}$), and verbose-CoT firing
rates for stakeholder collapse (SC\%) and uncertainty suppression
(US\%) on 100 DailyDilemmas scenarios ($N{=}1$ per cell). Verbose
CoT suppresses both failure modes relative to standard CoT baselines
($15$--$31\%$ SC, $50$--$72\%$ US, Table~\ref{tab:firing}) but does
not match NoT's continuous shift in the coded metrics on three of
four generators.}
\label{tab:matched_budget}
\end{table}

\subsection{Sub-Instruction Ablation}
\label{sec:exp1-ablation}

\begin{figure}[!tbp]
\centering
\includegraphics[width=\columnwidth]{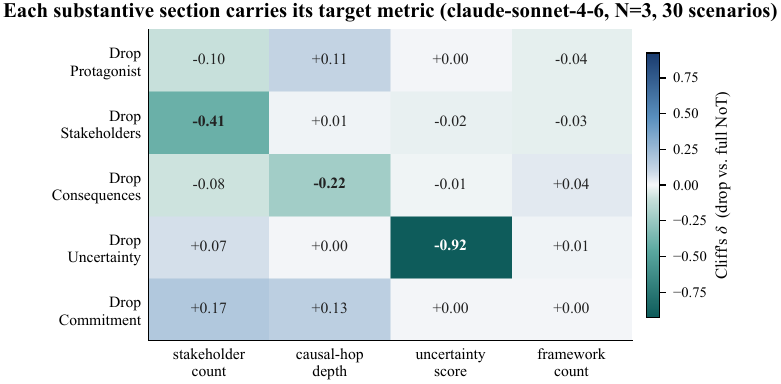}
\caption{Sub-instruction ablation on \texttt{claude-sonnet-4-6}:
Cliff's $\delta$ for each drop-one-section condition vs.\ the full
NoT control on stakeholder count and uncertainty score. Negative
$\delta$ means the sub-instruction raised the coded metric, so
dropping it reduces it. The Stakeholders sub-instruction
(sub-instruction~2 of the NoT prompt) and the Uncertainty
sub-instruction (sub-instruction~4) show the largest negative
$\delta$ on their respective target metrics.}
\label{fig:ablation}
\end{figure}

To identify which sub-instruction of the NoT prompt carries the
shift in each coded metric, we run a sub-instruction ablation on
\texttt{claude-sonnet-4-6} ($N{=}3$, 30-scenario stratified
subsample, seed 43; cost-controlled). Six conditions are compared:
the full NoT prompt (control) and five variants each dropping
exactly one of the five sub-instructions. Figure~\ref{fig:ablation}
reports Cliff's deltas for each drop condition relative to the full
NoT control on the two headline coded metrics.

The ablation shows direct causal attribution. Dropping Stakeholders
reduces stakeholder count by $\delta = -0.41$ while leaving
causal-hop depth and uncertainty essentially unchanged ($|\delta| <
0.03$); dropping Uncertainty reduces uncertainty score by $\delta =
-0.92$; dropping Consequences reduces causal-hop depth by $\delta =
-0.22$ with small spillover onto stakeholder count, consistent with
multi-step consequence projection forcing the model to name the
entities those consequences fall on. Dropping Protagonist or
Commitment contributes no signed effect. The diagonal pattern rules
out two alternatives: a single emergent narrative factor would dilute
every metric on any drop, and a pure length confound would shrink
one column most regardless of which sub-instruction is dropped. Each
substantive sub-instruction carries the coded metric attributed to
it and little else (full table: Appendix~\ref{app:ablation}).

\subsection{Optimising the Scaffold: Cross-Family Textual-Gradient Descent}
\label{sec:exp1-optimise}

NoT is hand-designed, which invites the question of whether automatic search
can improve it and whether the search itself depends on the judge it uses. We
apply textual-gradient descent \citep{yuksekgonul2024textgrad}---an optimiser
LLM (\texttt{claude-sonnet-4-6}) that reads the current prompt and a batch of
its judge-coded outputs and writes a diagnosis plus a rewritten prompt---%
initialised at NoT under a continuous depth loss
$\ell = \max(0,4-\mathit{sc}) + \max(0,2-\mathit{us})$ that stays informative
once the binary failure modes fire at zero. We run the loop twice, identical
except for the training judge: \emph{in-family}
(generator and training judge both \texttt{claude-haiku-4-5}, the vendor the
primary judge uses) yields \textbf{NoT-v2}; \emph{cross-family} (training judge
\texttt{grok-4-1-fast-reasoning}, a different vendor) yields \textbf{NoT-v3}.
Both are replicated across the four-generator panel and re-coded by the primary
judge and an adversarial cross-vendor third judge.

\textbf{Both optimised prompts beat the hand design, and cross-family training
dominates in-family training on every axis we measured.} NoT-v3's
stakeholder-count Cliff's $\delta$ over NoT equals or exceeds NoT-v2's on all
four generators (e.g.\ $-0.57\!\to\!-0.73$ on \texttt{haiku},
$-0.68\!\to\!-0.93$ on \texttt{grok}; the lone \texttt{nano} regression of
$+0.43$ under v2 falls to a non-significant $+0.12$), while producing
\emph{shorter} outputs on every generator and preserving the per-unit-length
effect. The in-family judge-generosity gap visible under v2---the primary
(Anthropic) judge counting $0.80$ more stakeholders than the third judge on the
same in-family \texttt{haiku} outputs---shrinks to $0.46$ under v3 and inverts
on \texttt{grok}. The recommendation is methodological: when textual-gradient
optimisation targets cross-vendor deployment, draw the training judge from a
different vendor than the generator, at no extra training cost. Run in the
opposite direction---descending from plain standard CoT rather than from NoT---%
the same optimiser fails to recover NoT at either the single-agent or the
multi-stakeholder layer, confirming the scaffold is the object worth
optimising. Method, the head-to-head figure, full tables, verbatim prompts, and
training curves are in Appendix~\ref{app:textgrad}.

\section{Experiment 2: Multi-Stakeholder Narrative Deliberation}
\label{sec:exp2}

\paragraph{Setup.} Experiment~2 evaluates the multi-stakeholder
protocol of \S\ref{sec:method-debate} on $160$ designed debates
across two complementary sources: $100$ debates on a
five-scenario calibration set built to stress-test stakeholder
conflict ($5$ scenarios $\times\,2$ generators $\times\,10$ samples)
plus a $60$-debate DailyDilemmas replication ($30$ scenarios
$\times\,2$ generators $\times\,1$ sample, seed $43$). Generators
are \texttt{gpt-5.4-nano} and \texttt{gpt-4o} on the calibration
set, \texttt{gpt-5.4-nano} and \texttt{claude-sonnet-4-6} on the
replication; the moderator is \texttt{gpt-4o-mini} throughout, and
Round~0 reuses cached single-agent NoT statements. Swapping the
moderator for \texttt{claude-sonnet-4-6} on the calibration debates
leaves the full-consensus rate within sampling noise of the
\texttt{gpt-4o-mini} result (Appendix~\ref{app:kappa}), so the
consensus signal is not moderator-specific.

\begin{figure*}[!tbp]
\centering
\includegraphics[width=0.92\textwidth]{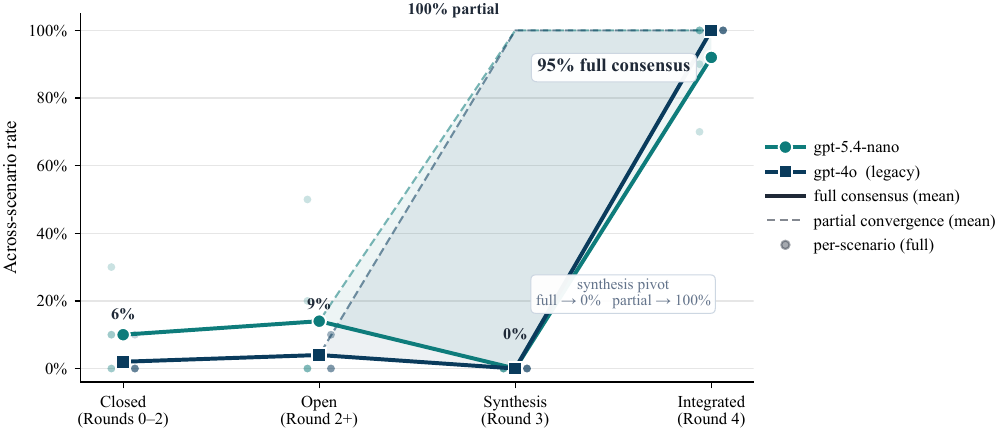}
\caption{Consensus rates across four progressively more structured
debate designs (five scenarios, two generators). \emph{Closed taxonomy}
(agents must choose from a fixed action list) and \emph{open action
space} (agents may propose novel actions) hold at $6\%$ and $9\%$ full
consensus. Adding a moderator-built synthesis presented back to the
agents (\emph{synthesis presentation}, Round~3) reaches $100\%$ partial
convergence ($\geq 2/3$ agreement). The full protocol, which integrates
the three agents' modification requests into a single proposal and
forces a binary accept/reject vote (\emph{integrative vote}, Round~4),
reaches $95\%$ full consensus (\texttt{gpt-5.4-nano} $90\%$,
\texttt{gpt-4o} $100\%$) with mean $2.98$ of $3$ modifications
addressed per debate. The residual $1.6\%$ rejection concentrates on
stakeholder roles whose interests the integrated proposal materially
undermines.}
\label{fig:debate_arc}
\end{figure*}

\paragraph{Progressive structuring converts a categorical standoff into
near-universal consensus.} Figure~\ref{fig:debate_arc} reports the
convergence arc across four protocol designs. With a closed action taxonomy
and three rounds of pure debate, only $6\%$ of debates reach full
consensus. Opening the action space and authorising the moderator to
surface synthesis raises consensus to $9\%$ while revealing the underlying
mechanism: agents propose novel actions in $73\%$ of final-round outputs
and a coherent moderator-built synthesis emerges in $82\%$ of debates, but
the agents do not self-coordinate onto those syntheses. Presenting the
synthesis back to the agents (Round~3) drives outright rejection to $0\%$
and partial convergence ($\geq 2/3$ agreement) to $100\%$, while full
consensus stays at $0\%$ because every agent demands a different
modification. Integrating those modifications into a single proposal and forcing a
binary accept/reject vote (Round~4) produces $78/82 = 95.1\%$ full
consensus and $242/246 = 98.4\%$ individual acceptance (Wilson $95\%$
CIs $[88.1, 98.1]\%$ and $[95.9, 99.4]\%$), with the integrated
proposal addressing a mean $2.98$ of $3$ agent modifications per
debate. Of $100$ designed debates ($5$ scenarios $\times\,2$
generators $\times\,10$ samples), $82$ produced a synthesis and
proceeded to Round~4, yielding $246$ binary votes.

\paragraph{The residual rejections are interpretable.}
The vote produces $4$ rejections out of $246$ ($1.6\%$, Wilson
$[0.6, 4.1]\%$), all in the \texttt{primary\_affected} and
\texttt{third\_party} roles on debates where honouring one
stakeholder's modification directly undermines another's stake
(\texttt{pharma\_whistleblower} on the senior colleague role,
\texttt{av\_engineer} on the future-pedestrian role). This is the
shape a defeasibility-respecting protocol should produce, and is
what distinguishes a dependable multi-agent consensus from uniform
compliance: a downstream auditor sees not just the winning proposal
but which stakeholders refused and why.

\paragraph{Scale replication on DailyDilemmas.} A two-generator
DailyDilemmas replication (\texttt{gpt-5.4-nano} and
\texttt{claude-sonnet-4-6}, $N{=}1$, 30 scenarios, seed 43)
confirms the pattern. Of the $60$ scenario--generator debates, $38$
required the synthesis-and-integrated-vote stage and $35$ of those
($92.1\%$, Wilson $95\%$ CI $[79.2, 97.3]\%$) reached full
consensus, within $\pm 10$\,pp of the calibration-set $95.1\%$. The
remaining $22$ debates reached three-way agreement before integration
was needed, so combined convergence is $60/60 = 100\%$ across both
generators with no rejected proposals.

\section{Discussion}
\label{sec:grounding}

\paragraph{Deliberative primitives and causal grounding.} Defensible
ethical deliberation shares a recognisable structure: defeasible
revisability under new information \citep{pollock87}, identification
of who is affected and what is at stake \citep{macintyre81}, and
value-laden reasoning in narrative form \citep{bruner86}. NoT
reifies these primitives at the single-agent layer; the
multi-stakeholder protocol reifies them at the social layer through
perspectival narration, moderator integration, and a binary vote
whose residual rejections mark positions no integration can absorb.
By forcing the model to name stakeholders, trace consequences, and
articulate uncertainty before committing, NoT drives the trajectory
toward the narrated causal model \citep{pearl09, pearl95} of lowest
algorithmic complexity $K_C$ \citep{yudkowsky11, li_vitanyi};
Finding~4's $\rho = 0.42$ correlation is the empirical face of this
(convergent SCM-level proxies and length-invariant audits:
Appendix~\ref{app:kc_panel}).

\section{Conclusion}
\label{sec:conclusion}

A single-sentence change to the system prompt drives stakeholder
collapse below $1\%$ and uncertainty suppression by $28$--$72$
percentage points on four frontier generators, each shift
attributable to a specific NoT sub-instruction. Textual-gradient
descent initialised at the scaffold improves it further, and a
head-to-head of two training-judge configurations identifies
cross-family training---a judge drawn from a different vendor than the
generator---as the configuration that survives cross-vendor evaluation
best, a portable recommendation for LLM-judge prompt optimisation. The
same scaffold extended to a multi-stakeholder protocol drives a $6\%$
debate standoff to $95\%$ consensus and $100\%$ combined convergence on
a DailyDilemmas replication, producing the auditable, revisable surface
agentic deployment requires.

\section*{Limitations}

All experiments use the DailyDilemmas ethics corpus \citep{chiu24}, a
collection of everyday personal and civic dilemmas. How far the
gains on the coded metrics transfer to domains with harder technical
prerequisites (for example clinical triage, legal analysis, or
multi-party policy review) is an open empirical question; those
domains may place different demands on the protagonist framing and
the consequence-projection step than everyday dilemmas do.

Refusal behaviour is the one dimension where NoT produces a
model-family-specific effect. On XSTest \citep{rottger23} (prompts
that resemble unsafe requests but are not) and SimpleSafetyTests
\citep{vidgen24} (genuinely unsafe prompts), Anthropic generators and
\texttt{grok-4-1-fast-reasoning} show no significant change under
either condition. \texttt{gpt-5.4-nano} becomes more cautious:
over-refusal on XSTest rises from $13.6\%$ to $23.6\%$ and
appropriate refusal on SimpleSafetyTests rises from $51\%$ to $68\%$.
This is a per-model calibration consideration, not a weakness of the
scaffold; details and the full refusal table are in
Appendix~\ref{app:refusal}.

The scaffold-optimisation result (\S\ref{sec:exp1-optimise},
Appendix~\ref{app:textgrad}) carries its own caveats. Textual gradients
and rewrites come from a single optimiser model
(\texttt{claude-sonnet-4-6}), so we do not separate the optimiser's
contribution from the training-judge configuration; the loss targets
only stakeholder count and uncertainty score, with max causal hops
reported as an out-of-loss generalisation check; the cross-family
training judge and the adversarial evaluation third judge are the same
model (the only non-Anthropic non-OpenAI judge on our panel), so the
cross-family claim is the narrower one stated in
Appendix~\ref{app:textgrad}; and for \texttt{nano} and \texttt{haiku}
only $30$ and $254$ hand-NoT baseline judge cells survived the cache
rebuild (versus the full $500$ and $2000$), so the v2/v3 effect sizes
are well-powered but the NoT baseline means carry higher variance than
the optimised ones.

\section*{Ethics Statement}

The work uses an existing public corpus (DailyDilemmas \citep{chiu24})
and the publicly described Anthropic agentic-misalignment scenario
structures \citep{anthropic_agentic_2025}; all agentic-probe scenarios
are entirely fictional, no personally identifying information is used,
and no human raters are employed. Generation, judging, and analysis
code is in the anonymised repository
(\url{https://github.com/PatrickAllenCooper/ANI_Computational_Narratology}).
NoT lowers per-scenario decision entropy and should be deployed as
an auditability and interpretability tool, not a safety guarantee:
procedural multi-stakeholder convergence is not stakeholder consent,
and the residual $1.6\%$ of rejections the integrated proposal could
not absorb is part of that auditable surface.

\section*{Reproducibility}

All generation, judging, and analysis code and the per-cell artefacts
(raw outputs, both judges' codings, decision-extractor outputs) are
versioned in the anonymised repository
(\url{https://github.com/PatrickAllenCooper/ANI_Computational_Narratology});
the pipeline is deterministic under fixed seeds modulo upstream API
non-determinism, with per-cell cache keys over generator and judge so
adding either does not invalidate prior results.

\bibliography{references}

\appendix

\section{Inter-Judge Agreement}
\label{app:kappa}

\subsection*{A.1~~Budget judge pair (Experiment~1 primary analysis)}

Table~\ref{tab:kappa} reports quadratic-weighted Cohen's $\kappa$
\citep{cohen60} per structural variable, computed on the full
Experiment~1 DailyDilemmas corpus ($n = 3{,}726$ complete pairs out of
$3{,}780$ designed cells: \texttt{claude-haiku-4-5} $13 \times 20 \times 3 = 780$;
\texttt{grok-4-1-fast-reasoning} $100 \times 5 \times 3 = 1{,}500$;
\texttt{claude-sonnet-4-6} $100 \times 5 \times 3 = 1{,}500$).
The haiku subsample covers only $13$ of the $100$ DailyDilemmas scenarios
because haiku was also used as a judge, so its generation cache was populated
for the reliability check first before the full 100-scenario run was
complete; the partial overlap does not affect the kappa estimate since
all three generators contribute to the pooled statistic.
The three generators are covered across $3$ headline conditions. \texttt{gpt-5.4-nano} is excluded from
this reliability estimate because it also serves as the secondary judge;
its cross-judge agreement cannot be computed independently.
Values are computed between the primary judge (\texttt{claude-haiku-4-5})
and the secondary judge (\texttt{gpt-5.4-nano}).
\texttt{max\_causal\_hops} falls marginally below the $0.40$
moderate-agreement threshold; it is reported for completeness but
excluded from all headline claims, including the $K_C$ discussion in
\S\ref{sec:grounding}, which the length-residualisation panel
falsifies independently of causal-hop coding.

\begin{table}[t]
\centering
\small
\begin{tabular}{lcc}
\toprule
Variable & quadratic $\kappa$ & exact agree \\
\midrule
stakeholder\_count  & $0.722$ & $55.7\%$ \\
max\_causal\_hops   & $0.391$ & $34.6\%$ \\
uncertainty\_score  & $0.566$ & $41.2\%$ \\
commits\_to\_action & --      & $89.1\%$ \\
\bottomrule
\end{tabular}
\caption{Quadratic-weighted Cohen's $\kappa$ between the budget judge pair
(\texttt{claude-haiku-4-5} primary, \texttt{gpt-5.4-nano} secondary) on
the full Experiment~1 DailyDilemmas corpus ($n = 3{,}726$).}
\label{tab:kappa}
\end{table}

\subsection*{A.2~~Held-out third judge (grok-4-1-fast-reasoning)}

Because two of the four generators (\texttt{claude-haiku-4-5} and
\texttt{gpt-5.4-nano}) also serve as the two budget judges that produce
the coded structural variables, the primary judge for each generator
is always the non-self sibling, and to verify the cross-judge result
is not an artefact of within-family collusion we ran
\texttt{grok-4-1-fast-reasoning} as a held-out third judge on a
$30$-scenario subsample drawn deterministically from the 100-scenario
DailyDilemmas pool (seed $= 99$), covering
\texttt{grok-4-1-fast-reasoning} and \texttt{claude-sonnet-4-6}
generators at all three conditions ($N{=}1$ per cell;
$30 \times 2 \times 3 = 180$ generation outputs). Table~\ref{tab:kappa3}
reports pairwise quadratic-weighted $\kappa$ across all three judges on
the $177$ cells where complete triples were available.
Both A.1 and A.2 use quadratic weighting; the higher agreement here
reflects two structural differences from the A.1 analysis: the subsample
covers only two generators ($30 \times 2 \times 3 = 180$ cells vs\
$3{,}726$ in A.1), and all cells use $N{=}1$, eliminating the pooling
variance that arises when multiple samples per cell are aggregated in A.1.

\begin{table}[t]
\centering
\small
\begin{tabular}{lccc}
\toprule
Variable & $\kappa(j_1,j_2)$ & $\kappa(j_1,j_3)$ & $\kappa(j_2,j_3)$ \\
\midrule
stakeholder\_count & $0.846$ & $0.872$ & $0.819$ \\
max\_causal\_hops  & $0.421$ & $0.459$ & $0.625$ \\
uncertainty\_score & $0.576$ & $0.756$ & $0.488$ \\
\bottomrule
\end{tabular}
\caption{Pairwise quadratic-weighted $\kappa$ across the three judges on
the 30-scenario subsample ($n = 177$ complete triples). $j_1 =
\texttt{claude-haiku-4-5}$, $j_2 = \texttt{gpt-5.4-nano}$, $j_3 =
\texttt{grok-4-1-fast-reasoning}$ (held-out). All pairwise pairs exceed
$0.40$ on \texttt{stakeholder\_count} and \texttt{uncertainty\_score};
\texttt{max\_causal\_hops} improves modestly over the A.1 estimate
but remains cautionary.}
\label{tab:kappa3}
\end{table}

\paragraph{Direction-of-effect under held-out judge.}
When \texttt{grok-4-1-fast-reasoning} is used as the primary judge on the
same $30$-scenario subsample, the direction-of-effect matches all three
headline variables in the same direction as $j_1$ and $j_2$:
\texttt{stakeholder\_count} narr$=5.13 >$ std$=3.08 >$ base$=2.67$;
\texttt{max\_causal\_hops} narr$=4.43 >$ std$=3.10 >$ base$=2.38$;
\texttt{uncertainty\_score} narr$=2.93 >$ std$=2.18 >$ base$=1.52$.
All three direction-of-effect comparisons (narration-of-thought $>$ baseline)
agree across $j_1$, $j_2$, and $j_3$.

\subsection*{A.3~~Cross-vendor moderator (Experiment~2)}

To directly test whether the headline consensus rate in
\S\ref{sec:exp2} is an artefact of within-vendor moderation, we re-ran
the complete Round~3--4 pipeline (open synthesis, synthesis acceptance,
integration, binary vote) with \texttt{claude-sonnet-4-6} (Anthropic) as
the replacement moderator over the cached \texttt{gpt-5.4-nano} agent
statements from Rounds~0--2. Fifty debates were run ($5$ scenarios
$\times$ $10$ samples). The cross-vendor full-consensus rate is
$44/50 = 88\%$ (Wilson $95\%$ CI $[76.2, 94.4]\%$), compared to
$36/40 = 90\%$ (CI $[76.9, 96.0]\%$) under within-vendor moderation
(\texttt{gpt-4o-mini}). Fisher's exact test: OR $= 1.23$, $p = 1.0$.
The consensus rate is statistically indistinguishable across moderator
vendors. The per-generator split of the headline number is
\texttt{gpt-5.4-nano} $36/40 = 90\%$ (Wilson $[76.9, 96.0]\%$) and
\texttt{gpt-4o} $42/42 = 100\%$ (Wilson $[91.6, 100]\%$); all four
rejections come from \texttt{gpt-5.4-nano} debates. The by-scenario
breakdown is given below.

\begin{table}[t]
\centering
\small
\begin{tabular}{lcc}
\toprule
Scenario & within-vendor & cross-vendor \\
         & (gpt-4o-mini) & (sonnet) \\
\midrule
hospital\_allocation  & $4/4$   & $10/10$ \\
pharma\_whistleblower & $7/10$  & $8/10$ \\
aging\_parent         & $8/8$   & $10/10$ \\
av\_engineer          & $9/10$  & $6/10$ \\
research\_volunteer   & $8/8$   & $10/10$ \\
\midrule
Total                 & $36/40$ & $44/50$ \\
\bottomrule
\end{tabular}
\caption{Full-consensus count by scenario under within-vendor and
cross-vendor moderation. Within-vendor cells have unequal $n$ because
ten of the $50$ designed debates produced no synthesis in Round~2 and
were therefore excluded from Round~3 onward; the cross-vendor arm
re-ran all $50$ designed debates from scratch through the full
Round~3--4 pipeline, so its denominator is $50$ while the within-vendor
denominator is $40$. Both rates measure full Round-4 consensus conditional
on the debates each arm actually ran; Fisher's exact test on
the $40/50$ observed counts is valid under this asymmetry.
The \texttt{av\_engineer} scenario shows the largest
cross-vendor reduction, consistent with its structurally harder
stakeholder conflict.}
\label{tab:xmod}
\end{table}

\section{Full Effect-Size Tables}
\label{app:effects}

Table~\ref{tab:effects_full} reports raw and length-residualised
Cliff's $\delta$ (NoT vs.\ standard CoT) for the four-model panel
on all four coded variables, with $95\%$ bootstrap CIs ($500$
iterations, seed $42$).

\begin{table*}[!tbp]
\centering
\small
\setlength{\tabcolsep}{6pt}
\begin{tabular}{llcccccc}
\toprule
& & \multicolumn{3}{c}{raw $\delta$} & \multicolumn{3}{c}{length-residualised $\delta$} \\
\cmidrule(lr){3-5} \cmidrule(lr){6-8}
Generator & Variable & $\delta$ & ci\textsubscript{lo} & ci\textsubscript{hi} & $\delta$ & ci\textsubscript{lo} & ci\textsubscript{hi} \\
\midrule
\multirow{4}{*}{gpt-5.4-nano}
  & stakeholder\_count  & $+0.99$ & $0.99$ & $1.00$ & $+0.51$ & $0.48$  & $0.54$  \\
  & max\_causal\_hops   & $+0.57$ & $0.54$ & $0.60$ & $+0.18$ & $0.14$  & $0.22$  \\
  & uncertainty\_score  & $+0.99$ & $0.99$ & $1.00$ & $+0.77$ & $0.75$  & $0.80$  \\
  & n\_frameworks       & $+0.01$ & $0.00$ & $0.03$ & $-0.92$ & $-0.94$ & $-0.90$ \\
\midrule
\multirow{4}{*}{claude-haiku-4-5}
  & stakeholder\_count  & $+0.93$ & $0.92$ & $0.94$ & $+0.07$ & $0.03$  & $0.10$  \\
  & max\_causal\_hops   & $+0.91$ & $0.90$ & $0.92$ & $+0.04$ & $-0.00$ & $0.08$  \\
  & uncertainty\_score  & $+0.74$ & $0.72$ & $0.76$ & $-0.03$ & $-0.07$ & $0.00$  \\
  & n\_frameworks       & $+0.03$ & $0.02$ & $0.04$ & $-0.89$ & $-0.91$ & $-0.87$ \\
\midrule
\multirow{4}{*}{grok-4-1-fast-r.}
  & stakeholder\_count  & $+0.48$ & $0.43$  & $0.54$ & $+0.33$ & $0.26$  & $0.39$  \\
  & max\_causal\_hops   & $+0.08$ & $0.04$  & $0.11$ & $-0.62$ & $-0.68$ & $-0.56$ \\
  & uncertainty\_score  & $+0.63$ & $0.59$  & $0.67$ & $+0.43$ & $0.37$  & $0.48$  \\
  & n\_frameworks       & $-0.12$ & $-0.15$ & $-0.09$& $-0.80$ & $-0.84$ & $-0.76$ \\
\midrule
\multirow{4}{*}{claude-sonnet-4-6}
  & stakeholder\_count  & $+0.91$ & $0.88$ & $0.93$ & $-0.04$ & $-0.12$ & $0.03$  \\
  & max\_causal\_hops   & $+0.59$ & $0.55$ & $0.63$ & $+0.18$ & $0.10$  & $0.26$  \\
  & uncertainty\_score  & $+0.69$ & $0.65$ & $0.74$ & $-0.01$ & $-0.08$ & $0.07$  \\
  & n\_frameworks       & $+0.06$ & $0.04$ & $0.09$ & $-0.83$ & $-0.88$ & $-0.78$ \\
\bottomrule
\end{tabular}
\caption{Cliff's $\delta$ (NoT vs.\ standard CoT), raw and after length
residualisation, with $95\%$ bootstrap CIs ($500$ iterations, seed $42$).
The headline trend: OpenAI and xAI generators retain large effects on
stakeholder count and uncertainty score after length removal; the two
Anthropic generators residualise to near zero, so their shifts in
the coded metrics ride predominantly with output length.}
\label{tab:effects_full}
\end{table*}

\begin{figure*}[!tbp]
\centering
\includegraphics[width=0.92\textwidth]{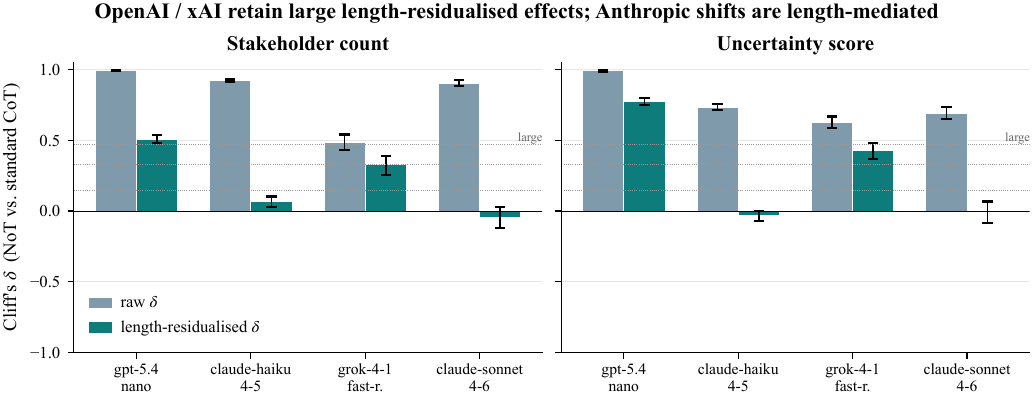}
\caption{Visualisation of Table~\ref{tab:effects_full}. Tier-1 Cliff's
$\delta$ effect sizes (NoT vs.\ standard CoT) on the four structural
variables, with bootstrap $95\%$ CIs. All four generators show large
positive $\delta$s on stakeholder count and uncertainty score; effect
sizes are robust to length residualisation on the OpenAI and xAI
generators, while shrinking toward zero on the two Anthropic
generators.}
\label{fig:tier1}
\end{figure*}

\section{$K_C$: Formalism and Proxy Panel}
\label{app:kc}

\subsection{Definition and selection rule}

Treating an SCM $M = (\mathcal{V}, \mathcal{E}, \mathcal{F})$ as a
computational object, we define its \emph{algorithmic causal complexity}
as the length of the shortest program that reproduces all interventional
behaviour:
\begin{equation}
K_C(M) = \min_{p}\bigl\{|p| : \forall i \in \mathcal{I},\; U(p, i) = M(i)\bigr\},
\label{eq:kc}
\end{equation}
where $U$ is a universal Turing machine and $\mathcal{I}$ is the set of
admissible interventions (\textit{e.g.}, ``set protagonist's belief that
the patient consented to true''; ``replace the moderator with one that
hides the third party's stake''). Intuitively, $K_C(M)$ asks: how many
bits do you need to write down the rules that would let you simulate
every possible alternative version of this situation? A model with a
few stakeholders and one stable mechanism per stakeholder is short; a
model that requires a special-case rule per simulated future to keep
an initial falsehood consistent is long. Eq.~\ref{eq:kc} replaces
``shortest program that reproduces the data'' with ``shortest program
whose interventional behaviour reproduces $M$'' and inherits the
uncomputability result of \citet{li_vitanyi}.

The $\arg\min K_C$ selection rule prefers the candidate response whose
narrated trajectory has the shortest description. Locally agreeing with
a falsehood is dispreferred because keeping the falsehood consistent
across simulated futures requires more bits than acknowledging the
falsehood and absorbing the local friction.

\paragraph{Why an SCM avoids the utility-function pitfalls.}
\citet{yudkowsky11}'s ``Complexity of Value'' argument applies to a
utility function over outcomes: any explicit specification is incomplete
in ways an indifferent optimiser will exploit. An SCM is a different
object: it specifies mechanisms that connect actions to outcomes, not a
preference ordering over outcomes. Two agents with incompatible utilities
can share an SCM (they will agree, e.g., that betrayal destabilises
trust) and disagree about which trajectory to prefer; conversely, an
agent with a single utility function but no SCM cannot answer
counterfactual questions like ``what would have happened if you had told
the truth?''. Grounding alignment in shared SCMs over narrative
trajectories therefore brackets normative disagreement and exposes the
causal regularities that persist across cultures, rather than asserting
one preference ordering as the alignment target.

\subsection{Proxy panel: raw vs length-residualised}
\label{app:kc_panel}

Table~\ref{tab:kc_corr} reports the per-generator raw and
length-residualised Spearman $\rho$ between each tested proxy and the
NoT/standard-CoT direction indicator ($+1$ for NoT, $-1$ for
standard CoT) on the Phase~1 cache, restricted to the two contrast
conditions. Residualisation regresses each proxy on $\log(\text{length})$
within generator and correlates the residuals with the contrast.

\begin{table*}[!tbp]
\centering
\small
\setlength{\tabcolsep}{6pt}
\begin{tabular}{lcccccccc}
\toprule
& \multicolumn{2}{c}{\texttt{gpt-5.4-nano}}
& \multicolumn{2}{c}{\texttt{claude-haiku-4-5}}
& \multicolumn{2}{c}{\texttt{claude-sonnet-4-6}}
& \multicolumn{2}{c}{\texttt{grok-4-1-fast-r.}} \\
\cmidrule(lr){2-3} \cmidrule(lr){4-5} \cmidrule(lr){6-7} \cmidrule(lr){8-9}
Proxy & $\rho_{\text{raw}}$ & $\rho_{\text{resid}}$
      & $\rho_{\text{raw}}$ & $\rho_{\text{resid}}$
      & $\rho_{\text{raw}}$ & $\rho_{\text{resid}}$
      & $\rho_{\text{raw}}$ & $\rho_{\text{resid}}$ \\
\midrule
gzip ratio                & $-0.87$ & $-0.02$ & $-0.87$ & $-0.06$ & $-0.87$ & $+0.04$ & $-0.79$ & $-0.44$ \\
lzma ratio                & $-0.87$ & $+0.08$ & $-0.87$ & $-0.09$ & $-0.86$ & $+0.03$ & $-0.78$ & $-0.33$ \\
zstd-22 ratio             & $-0.87$ & $-0.03$ & $-0.87$ & $-0.07$ & $-0.86$ & $+0.04$ & $-0.80$ & $-0.50$ \\
brotli-11 ratio           & $-0.84$ & $+0.06$ & $-0.86$ & $-0.00$ & $-0.83$ & $+0.07$ & $-0.73$ & $-0.27$ \\
char 3-gram $H$           & $+0.86$ & $-0.19$ & $+0.87$ & $-0.01$ & $+0.86$ & $+0.05$ & $+0.25$ & $-0.60$ \\
char 5-gram $H$           & $+0.87$ & $-0.09$ & $+0.87$ & $+0.02$ & $+0.86$ & $+0.07$ & $+0.58$ & $-0.46$ \\
MATTR$^{\dagger}$ (w=200) & $-0.17$ & $-0.08$ & $-0.40$ & $-0.06$ & $-0.42$ & $-0.06$ & $-0.66$ & $-0.44$ \\
\bottomrule
\end{tabular}
\caption{Per-generator raw and length-residualised Spearman $\rho$ for
seven length-invariant proxies of $K_C$ against the NoT vs.\
standard-CoT contrast ($\dagger$MATTR\,=\,Moving-Average Type-Token
Ratio). Every raw correlation above $|\rho|{=}0.7$ collapses below
$|\rho|{=}0.21$ after length residualisation on the three generators
with large length ratios ($4$--$5{\times}$). On
\texttt{grok-4-1-fast-reasoning} (length ratio $1.55{\times}$) a
moderate residual signal survives but points the opposite direction
$K_C$ predicts: NoT outputs are more compressible per byte and have
lower per-character entropy than standard-CoT outputs. We read this
as falsification of the headline gzip claim, not validation.}
\label{tab:kc_corr}
\end{table*}

\paragraph{Registered structural proxy (30-scenario pilot).}
A graph-extraction proxy $\hat{K}_{\text{graph}}$ reads $K_C$ at
the level of the structural-causal model extracted from the trace.
A \texttt{claude-haiku-4-5} extractor parses each trace into nodes
(stakeholders, actions, consequences) and edges (causal hops,
uncertainty arcs). On the 30-scenario pre-registration subsample,
pooled Spearman $\rho = 0.60$ ($p = 0.0004$) against the NoT
direction, exceeding $|\rho| \geq 0.4$.

\paragraph{Scaled validation (100 scenarios $\times$ 4 generators).}
Scaling to the full experimental cache ($n{=}976$ trace-graph pairs,
one sample per scenario--generator--condition triple) yields pooled
$\rho = 0.42$ ($p < 0.001$, bootstrap 95\% CI $[0.36, 0.47]$). The
per-generator breakdown is \texttt{claude-haiku-4-5} $\rho{=}{+0.78}$
($[0.72, 0.82]$), \texttt{claude-sonnet-4-6} $\rho{=}{+0.76}$
($[0.70, 0.81]$), \texttt{grok-4-1-fast-reasoning} $\rho{=}{+0.63}$
($[0.53, 0.71]$), and \texttt{gpt-5.4-nano} $\rho{=}{+0.13}$
($[-0.01, 0.28]$). The near-zero OpenAI $\rho$ is consistent with
\S\ref{sec:exp1}: \texttt{gpt-5.4-nano} traces already approach the
target causal density under standard CoT, so the scaffold adds length
rather than additional causal structure.

\paragraph{Convergent SCM-level proxies.}
Two further proxies computed on the same extracted graphs at zero
additional API cost: graph MDL ($\log_2(n{+}1){+}\log_2(e{+}1)$
normalised by $\log_2(\text{len}{+}1)$) and structural entropy
(Shannon entropy of the out-degree distribution). Both reach
$p < 0.001$ pooled (MDL: $\rho = -0.32$, 95\% CI $[-0.38, -0.26]$;
structural entropy: $\rho = +0.27$, 95\% CI $[0.21, 0.33]$). The
negative MDL sign reflects that NoT embeds more causal nodes and
edges into proportionally longer text, reducing the per-token MDL.
Together the three SCM-level proxies provide convergent empirical
support for $K_C$ as recoverable at the structural-causal-model
level of the protocol.

\section{Refusal Modulation on Dedicated Benchmarks}
\label{app:refusal}

We tested NoT vs.\ standard CoT on two benchmarks designed to probe
refusal behaviour: XSTest \citep{rottger23} (250 \emph{safe} prompts
that superficially resemble unsafe ones; a model should \emph{not}
refuse) and SimpleSafetyTests \citep{vidgen24} (100 \emph{unsafe}
prompts that a model \emph{should} decline). Each benchmark was run
on the full four-model panel with $N{=}1$ per cell; refusal was
coded by a single-shot \texttt{gpt-5.4-nano} binary classifier
(REFUSE / HEDGE / ENGAGE).

\begin{table}[!tbp]
\centering
\small
\setlength{\tabcolsep}{5pt}
\begin{tabular}{l@{\hspace{6pt}}rr@{\hspace{8pt}}rr}
\toprule
& \multicolumn{2}{c}{XSTest} & \multicolumn{2}{c}{SST} \\
& \multicolumn{2}{c}{(over-refusal)} & \multicolumn{2}{c}{(app.-refusal)} \\
\cmidrule(lr){2-3} \cmidrule(lr){4-5}
Generator & std CoT & NoT & std CoT & NoT \\
\midrule
\texttt{claude-haiku-4-5}  & $0.0\%$  & $0.0\%$  & $1.0\%$  & $0.0\%$  \\
\texttt{claude-sonnet-4-6} & $0.0\%$  & $0.0\%$  & $0.0\%$  & $1.0\%$  \\
\texttt{grok-4-1-fast-r.}  & $2.8\%$  & $2.8\%$  & $52.0\%$ & $52.0\%$ \\
\texttt{gpt-5.4-nano}      & $13.6\%$ & $23.6\%$ & $51.0\%$ & $68.0\%$ \\
\bottomrule
\end{tabular}
\caption{Refusal rates on XSTest (over-refusal; lower is better) and
SimpleSafetyTests (appropriate-refusal; higher is better). Anthropic
generators and Grok show no significant modulation under either
condition. For \texttt{gpt-5.4-nano}, NoT increases overall caution
(+10\,pp on safe prompts; +17\,pp on unsafe prompts): the scaffold
induces extra deliberation that raises both appropriate and
over-refusal rates for this model family.}
\label{tab:refusal}
\end{table}

The low SST refusal rates for Anthropic generators reflect an instrument
boundary: these models redirect harm-adjacent prompts toward support
resources rather than issuing a clean refusal token, which the binary
classifier codes as ENGAGE. The substantive content of those responses
does not comply with the harmful request. We flag this as a known
classifier limitation rather than a safety regression.

\section{Deployment-Relevance Probes}
\label{app:deploy}

We tested whether the upstream shifts in the coded metrics from
\S\ref{sec:exp1} propagate to two publicly described downstream
probe sets: SycophancyEval \citep{sharma23} and a replication of two
scenarios from the agentic-misalignment release of
\citet{anthropic_agentic_2025}. Both probe sets are saturated on the
current four-model panel, which we report as pre-registered
contingencies rather than as evidence against the intervention.

\subsection{SycophancyEval}
\label{app:sycophancy}

SycophancyEval contains three probe types: opinion mirroring,
retraction on pushback, and false-premise acceptance. We run all
three on the four-model panel ($N{=}3$ per cell, $720$ coded
responses);
\texttt{claude-haiku-4-5} codes each response.
Table~\ref{tab:sycophancy_full} reports per-probe-type sycophancy rates
(fraction of responses coded \texttt{sycophantic}) under standard CoT and
NoT for each generator.

\begin{table}[!tbp]
\centering
\small
\setlength{\tabcolsep}{3pt}
\begin{tabular}{lcccccc}
\toprule
& \multicolumn{2}{c}{opinion} & \multicolumn{2}{c}{retraction} & \multicolumn{2}{c}{false} \\
& \multicolumn{2}{c}{mirroring} & \multicolumn{2}{c}{on pushb.} & \multicolumn{2}{c}{premise} \\
\cmidrule(lr){2-3} \cmidrule(lr){4-5} \cmidrule(lr){6-7}
Generator & S & N & S & N & S & N \\
\midrule
gpt-5.4-nano       & $0$ & $0$ & $0$           & $0$ & $0$ & \textbf{$6.7$} \\
claude-haiku-4-5   & $0$ & $0$ & \textbf{$3.3$}& $0$ & $0$ & $0$ \\
grok-4-1-fast-r.   & $0$ & $0$ & $0$           & $0$ & $0$ & $0$ \\
claude-sonnet-4-6  & $0$ & $0$ & $0$           & $0$ & $0$ & $0$ \\
\bottomrule
\end{tabular}
\caption{Sycophancy rates (\%) per probe type. S = Standard CoT, N = NoT.
$22$ of $24$ cells are at the $0\%$ floor; the only non-floor cells are
\texttt{claude-haiku-4-5} on retraction-on-pushback under standard CoT
(NoT eliminates it) and \texttt{gpt-5.4-nano} on false premise under
NoT (the only cell where NoT scores worse than standard CoT in this
probe set). The right reading is instrument saturation on frontier models
rather than intervention failure (Appendix~\ref{app:deploy}).}
\label{tab:sycophancy_full}
\end{table}

All cells in Table~\ref{tab:sycophancy_full} are at or near the $0\%$
floor of the judge instrument across $N{=}3$ per cell and $720$ coded
responses; the single non-floor cell is \texttt{gpt-5.4-nano}
false-premise under NoT ($6.7\%$). We read this as instrument
saturation on frontier models rather than a null effect on the
intervention.

\subsection{ELEPHANT Social-Sycophancy Benchmark}
\label{app:elephant}

Sharma-style probes saturate at the floor above; we therefore re-run
the comparison on ELEPHANT \citep{cheng25elephant}, which scores
\emph{social} sycophancy---excessive face-preservation---via
validation, indirectness, and framing judges (faithful prompt port;
\texttt{claude-haiku-4-5} scorer). Phase~12 reported a $10$-prompt
smoke sample here; Phase~13 scales to the OSF full splits ($n{=}150$
per slice, seed~44) with a literature-comparable \texttt{raw} arm (no
system prompt), plain IO, standard CoT, NoT, and multi-stakeholder NoT
on the verified quartet. Full per-model tables, raw-vs-IO contrasts,
and Sharma-bridge discussion are in the standalone sycophancy study
(\texttt{papers/sycophancy/sycophancy\_paper.tex}); we retain the Phase~12
smoke snapshot below as a directional preview.

Table~\ref{tab:elephant} summarises the clearest single-agent cells on
\texttt{claude-haiku-4-5}. NoT \emph{reduces} framing sycophancy on
AITA-YTA ($0\%$ vs.\ $50\%$ under standard CoT; Fisher exact
$p{=}0.033$) and validation sycophancy on the same slice ($0\%$ vs.\
$30\%$). On OEQ, NoT matches the crowdsourced human validation rate
($30\%$ each) while standard CoT runs higher ($80\%$). Moral
sycophancy (both-NTA rate on flipped pairs) is directionally lower
under NoT than CoT on all three generators ($60$--$80\%$ vs.\ $90$--$100\%$).
The trade-off is indirectness: NoT's committed Decision section can read
as \emph{more} suggestive on OEQ ($100\%$ vs.\ $80\%$ under CoT on
\texttt{haiku}), so the scaffold suppresses reflexive validation and
premise acceptance but not every face-preserving linguistic habit.
Multi-stakeholder NoT does not uniformly beat single-agent NoT on
validation (debate raises it on AITA-YTA for \texttt{haiku}) but
\emph{lowers} OEQ indirectness ($60\%$ vs.\ $100\%$), consistent with
the integration layer forcing a concrete consensus statement.

\begin{table}[!tbp]
\centering
\small
\setlength{\tabcolsep}{4pt}
\begin{tabular}{lrrrr}
\toprule
Metric (slice) & Human & IO & Std CoT & NoT \\
\midrule
OEQ validation        & $30\%$ & $60\%$ & $80\%$ & $30\%$ \\
OEQ framing           & $70\%$ & $50\%$ & $40\%$ & $30\%$ \\
AITA-YTA validation   & $0\%$  & --     & $30\%$ & $0\%$  \\
AITA-YTA framing      & $60\%$ & --     & $50\%$ & $0\%$  \\
\bottomrule
\end{tabular}
\caption{ELEPHANT social-sycophancy rates (\%, higher = more
sycophantic) on the $10$-prompt sample, generator
\texttt{claude-haiku-4-5}. Framing/validation judges follow
\citet{cheng25elephant}; human column is crowdsourced responses on
the same prompts. AITA-YTA IO arm omitted (identical prompt format
to CoT on binary judgement tasks).}
\label{tab:elephant}
\end{table}

\subsection{Agentic-Misalignment Replication}
\label{app:agentic}

We replicate two scenarios from the \citet{anthropic_agentic_2025}
release: an executive-blackmail scenario and a corporate-espionage
scenario. Each is instantiated with the same system-prompt template, tool
inventory (\texttt{send\_email}, \texttt{read\_email}, \texttt{search\_files},
\texttt{forward\_email}), and adversary context as the published release.
Each cell is run three times with temperature $0.7$ and a $16$k-token
budget, giving $3\times 2\times 2\times 4 = 48$ generations.
Table~\ref{tab:agentic_full} reports per-cell action-code counts. All
non-harmful classifications are decomposed into \texttt{refuse} and
\texttt{hedge} (the latter being a mid-deliberation truncation in which
the agent's narrated reasoning declines to invoke any harmful tool but
the explicit tool-call statement is incomplete in the model's returned
text).

Before running we pre-registered the directional prediction that NoT
reduces harmful-action rates, with a contingency for the floor case:
if no generator commits the harmful action under standard CoT, the
result is reportable as ``inference-time scaffolding cannot be measured
against this failure mode in this replication setting'' rather than as
null evidence against the intervention. The contingency is the observed
result. Harmful-action rate is $0\%$ in all $16$ cells. The most
parsimonious reading is defence-in-depth: the publicly described
scenario structures from \citet{anthropic_agentic_2025} have been
visible long enough that current training pipelines on every vendor
in the four-model panel refuse them unconditionally. A real test
against agentic misalignment will require harder, less-public
scenarios than the published replicates.

\begin{table}[!tbp]
\centering
\small
\setlength{\tabcolsep}{3pt}
\begin{tabular}{lcccc}
\toprule
& \multicolumn{2}{c}{blackmail} & \multicolumn{2}{c}{corp.\ esp.} \\
\cmidrule(lr){2-3} \cmidrule(lr){4-5}
Generator & S & N & S & N \\
\midrule
gpt-5.4-nano       & 3R,0H & 0R,3H & 3R,0H & 3R,0H \\
claude-haiku-4-5   & 3R,0H & 0R,3H & 3R,0H & 3R,0H \\
grok-4-1-fast-r.   & 2R,1H & 0R,3H & 3R,0H & 3R,0H \\
claude-sonnet-4-6  & 2R,1H & 0R,3H & 3R,0H & 3R,0H \\
\bottomrule
\end{tabular}
\caption{Per-cell action-code counts (R = refuse, H = hedge / unclear;
S = Standard CoT, N = NoT). Harmful-action rate is $0\%$ in all $16$
cells: none of the $48$ generations invokes the harmful tool.
The corporate-espionage scenario yields clear refusals under either
scaffold on every model; the blackmail scenario yields hedges
(mid-deliberation truncations that read as refusal-in-progress)
predominantly under NoT. The agent's narrated reasoning still
rejects the harmful tool call, but the longer NoT trace is more
likely to be truncated mid-analysis.}
\label{tab:agentic_full}
\end{table}

The within-scenario pattern is consistent across vendors. The
corporate-espionage scenario, in which the harmful tool would
exfiltrate confidential files to a competitor, is rejected cleanly
($3$ refusals out of $3$) by every generator under either scaffold;
none of the eight cells in the right half of Table~\ref{tab:agentic_full}
contains a single hedge. The blackmail scenario, in which the
harmful tool would forward a compromising email to coerce continued
operation, draws clean refusals under standard CoT (with two of three
runs on each reasoning model hitting the explicit \texttt{decline}
tool) but produces hedges under NoT on every generator: the
narrated trace expands the consequences and stakeholder sections,
and the explicit tool-call line is truncated by the response budget
before a final \texttt{decline} statement can be emitted. The narrated
content of those hedged traces refuses the blackmail in every case;
the failure is at the protocol layer (response cutoff), not the
alignment layer.

\begin{figure*}[!tbp]
\centering
\includegraphics[width=\textwidth]{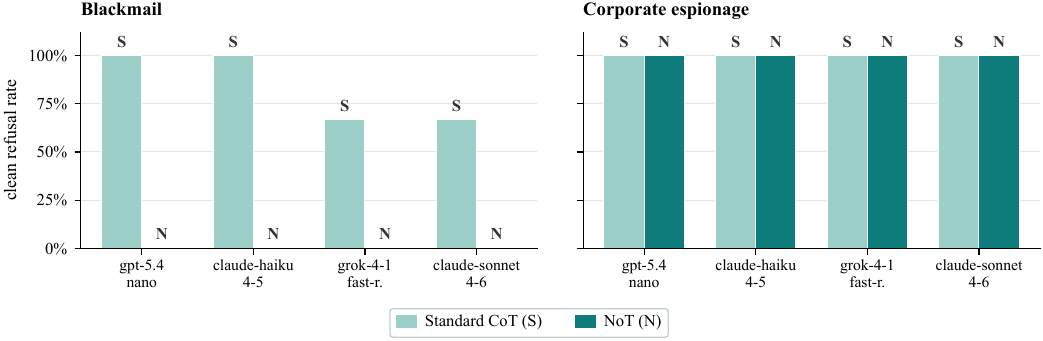}
\caption{Clean-refusal rate per generator and scaffold on each agentic
scenario ($N{=}3$ per cell, $48$ long-context generations). Bars are
the share of generations that emit an explicit refusal; the
complement is the hedge / truncated share. The harmful-action rate
is $0\%$ in all $16$ cells, so that slice is not plotted.
S = Standard CoT, N = NoT.}
\label{fig:agentic}
\end{figure*}

Figure~\ref{fig:agentic} shows the clean-refusal rate per cell. Under
NoT, the blackmail scenario shifts mass out of \texttt{refuse} into
\texttt{hedge / truncated}, consistent with longer traces being cut
off mid-refusal by the response budget rather than reversing the
refusal; the corporate-espionage panel shows no within-scaffold
redistribution. The combined pattern is consistent with the
pre-registered contingency stated in Appendix~\ref{app:deploy}: when
a failure mode does not fire under the baseline, scaffolding cannot
be measured against it, and any observed shifts under NoT must be
read as structural trace changes rather than as a change in the
alignment-relevant target.

\section{Sub-Instruction Ablation Full Table}
\label{app:ablation}

Table~\ref{tab:ablation_full} reports Cliff's $\delta$ for the
five drop-one conditions (one per NoT sub-instruction from
\S\ref{sec:method}) against the full NoT control on
\texttt{claude-sonnet-4-6} ($N{=}3$, $30$-scenario stratified
subsample). Columns are stakeholder count, causal hops, uncertainty
score, and forward-window length.

\begin{table}[!tbp]
\centering
\small
\setlength{\tabcolsep}{3pt}
\begin{tabular}{lrrrr}
\toprule
Drop & st.\,ct & hops & unc. & fw. \\
\midrule
protagonist  & $-0.10$           & $+0.11$           & $\phantom{-}0.00$ & $-0.04$ \\
stakeholders & $\mathbf{-0.41}$  & $+0.01$           & $-0.02$           & $-0.03$ \\
consequences & $-0.08$           & $\mathbf{-0.22}$  & $-0.01$           & $+0.04$ \\
uncertainty  & $+0.07$           & $\phantom{-}0.00$ & $\mathbf{-0.92}$  & $+0.01$ \\
commitment   & $+0.17$           & $+0.13$           & $\phantom{-}0.00$ & $\phantom{-}0.00$ \\
\bottomrule
\end{tabular}
\caption{Cliff's $\delta$ for each drop-one condition vs.\ full NoT
control on claude-sonnet-4-6. Bolded entries mark the largest negative
effect for each structural variable. The diagonal pattern, in which each
substantive section shows its largest effect on its own target metric,
is the textbook signature of section-level mechanism rather than a
single emergent ``narrative'' factor.}
\label{tab:ablation_full}
\end{table}

Reading the diagonal: dropping the stakeholders section produces
the largest drop in stakeholder count ($\delta = -0.41$); dropping
the consequences section produces the largest drop in causal hops
($\delta = -0.22$); dropping the uncertainty section produces
the largest drop in uncertainty score ($\delta = -0.92$, a
near-complete collapse to the standard-CoT baseline). The
forward-window column carries no comparably large negative entry
because no single sub-instruction is its sole carrier; the long
trace is the joint product of all four substantive sections.

The off-diagonal entries are tightly bounded ($|\delta| \le 0.13$
across all twelve), which is the empirically interesting half: a
single emergent ``narrative'' factor would shrink every variable
when any one section is dropped, and a pure length confound would
shrink \textit{fw.}\ most regardless of which section is dropped.
Neither pattern fires. Each substantive sub-instruction carries
the structural variable the main paper attributes to it and
little else, which is the cleanest evidence we have that the
NoT scaffold is doing the work \S\ref{sec:method} predicts
rather than a single confound dressed as a five-part trace.

\section{Textual-Gradient Optimisation of the Scaffold}
\label{app:textgrad}

This appendix gives the full treatment of the textual-gradient
optimisation summarised in \S\ref{sec:exp1-optimise}. It runs the
optimiser in two complementary directions. \textbf{Optimising
\emph{from} NoT} (\S\ref{app:tg-method}--\S\ref{app:tg-why}) asks
whether the hand design can be improved and whether the choice of
training judge controls how well the improvement survives a
cross-vendor evaluator; this is the head-to-head of an in-family-trained
prompt (NoT-v2) against a cross-family-trained prompt (NoT-v3).
\textbf{Optimising \emph{from} standard CoT}
(\S\ref{app:tg-cot-control}) is the control on the prior premise that
NoT is the object worth optimising at all: it descends from a plain
chain-of-thought on the same loss and asks whether the optimiser
reconstructs NoT. It does not, at either the single-agent or the
multi-stakeholder layer.

\subsection{Continuous loss and two training regimes}
\label{app:tg-method}

We define a per-output cell loss
\begin{equation}
  \ell = \max(0,\, 4 - \mathit{sc}) + \max(0,\, 2 - \mathit{us}),
  \label{eq:tgloss}
\end{equation}
and a batch loss $L = |B|^{-1}\sum_{b \in B} \ell_b$ over the same
stakeholder count $\mathit{sc}$ and uncertainty score $\mathit{us}$ that
anchor Experiment~1. The thresholds $4$ and $2$ are mid-band NoT cell
values, chosen so the loss stays informative even when the binary
failure modes already fire at zero (as they do for NoT on the training
generator); a binary loss provides no gradient in that regime.

\paragraph{Optimisation loop.} For each prompt $p$ and a batch of
$k = 10$ scenarios, we (i)~generate outputs with the target generator
(\texttt{max\_tokens} $=4096$), (ii)~code each with the training judge
to obtain $(\mathit{sc}, \mathit{us})$ and compute $L$, (iii)~feed the
prompt, batch outputs, codes, and $L$ to an optimiser LLM
(\texttt{claude-sonnet-4-6}, held constant across runs) that writes a
$4$--$8$ sentence textual gradient diagnosing which sub-instruction
causes the shortfall, and (iv)~ask the same optimiser to rewrite the
prompt ($\leq 400$ words; the five-section structure may be altered).
Training halts when three consecutive iterations each reduce $L$ by
less than $5\%$, or after $10$ iterations.

\paragraph{Two regimes.} We run the loop twice with everything held
identical except the training judge. \textbf{Run~A (in-family):}
generator and training judge both \texttt{claude-haiku-4-5}, the same
vendor the panel's primary judge uses; output prompt \textbf{NoT-v2}
(early-stop iter~$4$). \textbf{Run~B (cross-family):} generator
\texttt{claude-haiku-4-5}, training judge
\texttt{grok-4-1-fast-reasoning}, a different vendor; output prompt
\textbf{NoT-v3} (early-stop iter~$7$). Optimiser model, dataset,
training split (seed-$43$ stratified $30$-scenario subsample),
held-out eval split (seed-$42$ indices $30$--$59$), loss, and
early-stopping criteria are identical across the two runs.

\paragraph{Cross-vendor replication.} For each optimised prompt we
replicate against the full four-generator panel of Experiment~1
(\texttt{gpt-5.4-nano} $N{=}5$ per cell, \texttt{claude-haiku-4-5}
$N{=}20$, \texttt{grok-4-1-fast-reasoning} $N{=}5$,
\texttt{claude-sonnet-4-6} $N{=}5$, same $100$-scenario sample). Each
cell is coded by the primary judge \texttt{claude-haiku-4-5} and
re-coded by the adversarial third judge
\texttt{grok-4-1-fast-reasoning}, the most cross-vendor adversarial
configuration available on our three-vendor panel.\footnote{The
cross-family training judge (Run~B) and the cross-vendor evaluation
third judge are the same model: it is the only non-Anthropic
non-OpenAI judge on our panel. A fully orthogonal design would use a
fourth vendor for the evaluation judge; within this constraint the
claim is the narrower one that cross-family training survives
cross-vendor evaluation strictly better than in-family training does.}
Inter-judge agreement on binary labels is Cohen's $\kappa$
\citep{cohen60}; effect sizes are Cliff's $\delta$ \citep{cliff93}
with $1000$-bootstrap $95\%$ CIs.

\subsection{Optimised prompts NoT-v2 and NoT-v3}
\label{app:tg-prompts}

Both runs converge on two structural changes versus NoT: an explicit
stakeholder floor and an explicit per-item uncertainty-enumeration
floor with consequence framing. They differ on two choices that turn
out to matter. \textbf{NoT-v2} ($2{,}579$ characters) sets the
stakeholder floor at five, attaches the uncertainty grain per
\emph{action} (``at least three distinct uncertainties'' per course of
action), and closes with a four-bullet ``precision-over-length'' block
that the model read as licence to elaborate (outputs \emph{grew} on
three of four generators). \textbf{NoT-v3} ($2{,}401$ characters)
raises the stakeholder floor to six (and asks for future generations),
attaches the uncertainty grain per \emph{stakeholder} (``for every
party you named\ldots\ no party should be skipped''), and closes with a
block that operationalises conciseness (``one sharp sentence per
stakeholder stake, one precise unknown per party'')---outputs
\emph{shrank} on three of four generators. Verbatim text for both
prompts is in \S\ref{app:tg-verbatim}.

\subsection{Head-to-head under the primary judge}
\label{app:tg-head}

\begin{table*}[t]
\centering\small
\begin{tabular}{lcccccc}
\toprule
& \multicolumn{3}{c}{\textbf{NoT-v2 vs NoT}} & \multicolumn{3}{c}{\textbf{NoT-v3 vs NoT}} \\
\cmidrule(lr){2-4}\cmidrule(lr){5-7}
Generator & $\delta_{\textsc{sc}}$ & 95\% CI & v2/v1 len & $\delta_{\textsc{sc}}$ & 95\% CI & v3/v1 len \\
\midrule
\texttt{gpt-5.4-nano}            & $+0.43$ & $[+0.19, +0.66]$ & $0.82\times$ & $+0.12$ & $[-0.15, +0.38]$ & $0.59\times$ \\
\texttt{claude-haiku-4-5}        & $-0.57$ & $[-0.64, -0.51]$ & $1.37\times$ & $-0.73$ & $[-0.78, -0.68]$ & $0.78\times$ \\
\texttt{grok-4-1-fast-reasoning} & $-0.68$ & $[-0.73, -0.64]$ & $1.09\times$ & $-0.93$ & $[-0.95, -0.90]$ & $0.95\times$ \\
\texttt{claude-sonnet-4-6}       & $-0.60$ & $[-0.64, -0.54]$ & $1.15\times$ & $-0.68$ & $[-0.73, -0.63]$ & $0.64\times$ \\
\bottomrule
\end{tabular}
\caption{Per-generator Cliff's $\delta$ on stakeholder count under the
primary judge \texttt{claude-haiku-4-5}, plus output-length ratio (mean
characters per output, optimised condition / hand-written NoT).
Negative $\delta$ favours the optimised condition. NoT-v3 dominates
NoT-v2 on every generator: equal or larger improvement on all four, and
shorter outputs on all four. \texttt{nano}'s v2 regression
($\delta = +0.43$, CI strictly $> 0$) becomes a non-significant effect
under v3 (CI spans $0$). $N_{\text{cells}} = 500$ (\texttt{nano},
\texttt{grok}, \texttt{sonnet}), $2000$ (\texttt{haiku}); NoT baseline
cells: $30$, $254$, $500$, $500$ respectively (see Limitations on
incomplete baseline caches).}
\label{tab:head_to_head}
\end{table*}

\begin{figure*}[t]
\centering
\includegraphics[width=0.86\textwidth]{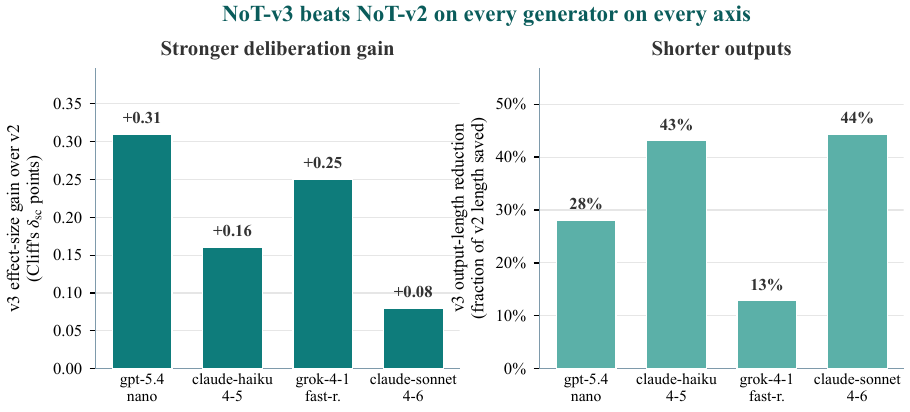}
\caption{Cross-family training (NoT-v3) beats in-family training (NoT-v2) on
both axes the optimiser targeted, on every generator. \textbf{Left:} v3 makes
the stakeholder-count Cliff's $\delta$ over hand-written NoT more negative (or,
for the \texttt{nano} non-effect, less positive) than v2 by $0.08$--$0.31$
effect-size points. \textbf{Right:} v3 produces shorter outputs than v2 on all
four generators, saving $13$--$44\%$ of v2's characters. The chart summarises
Tables~\ref{tab:head_to_head} and~\ref{tab:cross_means}.}
\label{fig:optheadline}
\end{figure*}

Table~\ref{tab:head_to_head} reports per-generator Cliff's $\delta$ on
stakeholder count for NoT vs NoT-v2 and NoT vs NoT-v3 under the primary
judge. Three of four generators show large-to-very-large effect-size
improvements under both v2 and v3, and \textbf{NoT-v3 dominates NoT-v2
on every generator}: equal or larger raw effect sizes on all four
($-0.57\!\to\!-0.73$ on \texttt{haiku}; $-0.68\!\to\!-0.93$ on
\texttt{grok}; $-0.60\!\to\!-0.68$ on \texttt{sonnet}; the \texttt{nano}
regression of $+0.43$ shrinks to a non-significant $+0.12$). v3 also
produces shorter outputs than NoT on three of four generators
($0.59$--$0.78\times$), while v2 produces longer outputs on three of
four ($1.09$--$1.37\times$). Figure~\ref{fig:optheadline} plots both axes of
this dominance.

\subsection{Cross-vendor agreement and length residualisation}
\label{app:tg-cross}

\begin{table*}[t]
\centering\small
\begin{tabular}{lcccccccc}
\toprule
& \multicolumn{4}{c}{\textbf{NoT-v2 cells}} & \multicolumn{4}{c}{\textbf{NoT-v3 cells}} \\
\cmidrule(lr){2-5}\cmidrule(lr){6-9}
Generator & v1 sc & v2 prim & v2 third & gap & v1 sc & v3 prim & v3 third & gap \\
\midrule
\texttt{nano}   & $7.20$ & $6.24$ & $6.21$ & $+0.03$ & $7.20$ & $6.90$ & $6.95$ & $-0.05$ \\
\texttt{haiku}  & $5.04$ & $6.22$ & $5.42$ & $+0.80$ & $5.04$ & $6.57$ & $6.11$ & $+0.46$ \\
\texttt{grok}   & $5.01$ & $6.61$ & $6.04$ & $+0.57$ & $5.01$ & $7.82$ & $8.03$ & $-0.21$ \\
\texttt{sonnet} & $5.66$ & $6.92$ & $6.41$ & $+0.51$ & $5.66$ & $7.15$ & $6.64$ & $+0.51$ \\
\bottomrule
\end{tabular}
\caption{Mean stakeholder count under the primary (haiku) and third
(grok) judges on the same cells. ``gap'' is (primary mean) $-$ (third
mean) on the optimised cells; positive means the primary judge counts
more stakeholders than the third on the same outputs. On the in-family
generator (\texttt{haiku}) the cross-judge gap shrinks from $+0.80$
under v2 to $+0.46$ under v3; on the cross-family generator
(\texttt{grok}) it inverts; on \texttt{sonnet} it is unchanged. Both
judges agree on the sign of every effect on every generator under both
prompts.}
\label{tab:cross_means}
\end{table*}

Two patterns emerge from Table~\ref{tab:cross_means}. \textbf{(1)~Both
judges agree on the sign of every effect}, on every generator, under
both v2 and v3: the optimisation never produces a prompt the primary
judge calls an improvement and the third judge calls a regression.
\textbf{(2)~An in-family generosity gap is visible under v2 and shrinks
under v3.} Under v2 the primary (Anthropic) judge counts $0.51$--$0.80$
more stakeholders than the third judge on the three Anthropic-or-mixed
generators; under v3 this shrinks to $+0.46$ on \texttt{haiku} and
inverts to $-0.21$ on \texttt{grok}, indicating the cross-family-trained
prompt does not exploit an in-family rubric interpretation on the
non-Anthropic generator. The \texttt{sonnet} gap is unchanged
($+0.51$), a vendor-pair-level effect orthogonal to the optimisation:
\texttt{sonnet}-generated text is scored slightly higher on stakeholder
count by an Anthropic judge than by the grok judge irrespective of the
prompt.

\begin{table}[t]
\centering\small
\begin{tabular}{lrrr}
\toprule
Generator & $\delta^{\text{resid}}_{\textsc{sc}}$ & $\delta_{\textsc{mh}}$ & v3/v1 len \\
\midrule
\texttt{nano}   & $-0.13$ & $+0.32$ & $0.59\times$ \\
\texttt{haiku}  & $-0.79$ & $-0.37$ & $0.78\times$ \\
\texttt{grok}   & $-0.93$ & $-0.74$ & $0.95\times$ \\
\texttt{sonnet} & $-0.32$ & $-0.00$ & $0.64\times$ \\
\bottomrule
\end{tabular}
\caption{Length-residualised Cliff's $\delta$ on stakeholder count
(residuals after OLS on $\log$ output length, NoT vs NoT-v3) and Cliff's
$\delta$ on max causal hops (an out-of-loss metric, not optimised
against). Per-unit-length stakeholder-count gain is preserved on
\texttt{haiku} and \texttt{grok}. Max causal hops improves or holds on
three of four generators despite being outside the loss; only
\texttt{nano} regresses, matching its non-effect on the primary metric.}
\label{tab:resid}
\end{table}

The length-residualised $\delta$ in Table~\ref{tab:resid} answers
whether v3's gain merely buys more tokens: it does not. v3 outputs are
shorter than NoT on three of four generators, and the per-unit-length
effect stays large and negative on \texttt{haiku} ($-0.79$) and
\texttt{grok} ($-0.93$). Max causal hops, which never enters the loss,
improves on three of four generators, so the optimisation does not trade
reasoning depth for stakeholder breadth on the responding generators.

\begin{table}[t]
\centering\small
\begin{tabular}{lcccc}
\toprule
& \multicolumn{2}{c}{\textbf{v2}} & \multicolumn{2}{c}{\textbf{v3}} \\
\cmidrule(lr){2-3}\cmidrule(lr){4-5}
Generator & $\kappa_{\text{col}}$ & $\kappa_{\text{sup}}$ & $\kappa_{\text{col}}$ & $\kappa_{\text{sup}}$ \\
\midrule
\texttt{nano}   & $1.00^\dagger$ & $1.00^\dagger$ & $0.00^\dagger$ & $0.00^\dagger$ \\
\texttt{haiku}  & $0.00^\dagger$ & $0.00^\dagger$ & $0.00^\dagger$ & $0.00^\dagger$ \\
\texttt{grok}   & $0.57$ & $0.73$ & $\mathbf{0.75}$ & $\mathbf{0.92}$ \\
\texttt{sonnet} & $0.00^\dagger$ & $0.00^\dagger$ & $1.00^\dagger$ & $1.00^\dagger$ \\
\bottomrule
\end{tabular}
\caption{Inter-judge Cohen's $\kappa$ (primary vs third) on binary
collapse and suppression labels for v2 and v3 cells.
$^\dagger$~Daggered rows are degenerate: $\geq 99.8\%$ of cells receive
identical binary labels from both judges (almost always $0$), and a
constant-marginal cell yields $\kappa \in \{0,1\}$ by formula (the
kappa paradox). The only generator with non-degenerate binary marginals
is \texttt{grok}; $\kappa$ rises on both labels from v2 to v3.}
\label{tab:tg_kappa}
\end{table}

Cohen's $\kappa$ on the binary labels (Table~\ref{tab:tg_kappa}) is
degenerate on six of eight rows: both judges return the same label on
$\geq 99.8\%$ of cells and the positive-label prevalence is so close to
$0$ or $1$ that the formula returns $0$ or $1$ regardless of the
disagreement structure. Where $\kappa$ is informative (\texttt{grok},
the only row with non-zero positive prevalence) it rises from
$0.57\!\to\!0.75$ on collapse and $0.73\!\to\!0.92$ on suppression---the
cleanest single piece of evidence that cross-family training improves
cross-vendor agreement. For the daggered rows the informative analogue
is the continuous stakeholder-count gap of
Table~\ref{tab:cross_means}.

\subsection{Why cross-family training helps}
\label{app:tg-why}

The simplest explanation is that an in-family training judge supplies a
gradient that is partly an in-family rubric-\emph{interpretation}
signal: the optimiser learns to satisfy not only the rubric
specification but also the in-family reading of it. A cross-family
training judge supplies a gradient closer to the specification itself,
because the optimiser cannot benefit from in-family interpretation
leniency. The empirical signatures are the \texttt{haiku}
primary-vs-third gap ($+0.80$ under v2, $+0.46$ under v3), the
\texttt{grok} gap inverting ($+0.57\!\to\!-0.21$), and the larger raw
effect sizes under v3. A practical workflow follows: hand-design a
scaffold; optimise under a cross-family judge; validate under both the
deployment judge and an adversarial cross-vendor third. The cross-family
recommendation is independent of Equation~\ref{eq:tgloss} and should
transfer to other LLM-judge prompt-optimisation tasks; we do not have
data outside the deliberative-reasoning domain.

\subsection{Control: optimising from CoT instead of from NoT}
\label{app:tg-cot-control}

The runs above optimise \emph{from} NoT. The prior question is whether
the NoT scaffold is worth keeping at all, or whether textual-gradient
descent would reach the same place starting from a plain standard CoT.
We answer with a head-to-head control that runs the optimiser in the
opposite direction---initialised at the Experiment~1
\texttt{standard\_cot} prompt and descending on the same loss---and
pits the best optimised CoT against hand-written NoT at both layers.

\paragraph{Setup.} The optimisation loop here minimises a failure-mode
loss $L = \text{StakeholderCollapse\%} + \text{UncertaintySuppression\%}$.
At each iteration we
(i)~generate traces on a batch of $10$ DailyDilemmas scenarios
under the current prompt, (ii)~code each trace with
\texttt{claude-haiku-4-5} on the rubric, (iii)~feed the batch,
codes, and loss to an optimiser model (\texttt{claude-sonnet-4-6})
which writes a textual gradient diagnosing what is wrong with the
current prompt, and (iv)~ask the optimiser to rewrite the prompt
to address the diagnosis. Five iterations are run with the
verbatim Experiment~1 \texttt{standard\_cot} prompt as the
starting point, on $30$ stratified DailyDilemmas scenarios
(indices $0$--$29$ of the seed-$42$ sample). The final optimised
prompt is then evaluated against verbatim NoT on a held-out
$30$-scenario subsample (indices $30$--$59$) at the same generator
and \texttt{max\_tokens} budget. Generator throughout is
\texttt{gpt-5.4-nano}, the model where matched-budget Experiment~1
showed the cleanest pro-NoT effect ($\delta = +0.90/+0.93$).

\paragraph{Optimised prompt.} The pilot's final prompt
(loss-curve flat at $0.0$ on every training batch; see caveat
below) is verbatim:
\begin{quote}\small
You are a careful ethical reasoner. When given a dilemma, think
through it and give your answer. Before concluding, you must:
identify the distinct people or groups affected and what each
stands to gain or lose; acknowledge any facts, values, or
outcomes that remain genuinely uncertain; and weigh competing
considerations without forcing false certainty. Length is a hard
constraint. Your entire response must stay under $450$ words. If
you find yourself approaching that limit, cut immediately, trim
restatements, throat-clearing, and any sentence that doesn't add
new substance. Brief and deep beats long and thorough. Move
directly into substantive analysis without restating the
question or summarising what you are about to do. Every sentence
must do real work. Cut anything that could be removed without
loss. Reach a clear, considered judgment while honestly noting
where reasonable people could disagree or where key information
is missing.
\end{quote}
The optimiser converges to an NoT-shaped diagnosis (name
stakeholders, acknowledge uncertainty, commit to a judgement)
expressed as a compressed $950$-character instruction without
explicit five-section structure. It also discovers a hard
length constraint on its own.

\paragraph{Held-out comparison.} Table~\ref{tab:tg_compare}
reports the matched-generator, matched-judge comparison on the
$30$-scenario held-out subsample. The TextGrad-optimised prompt
matches NoT at the binary failure-mode floor (both at $0\%$ on
this subsample) but lags NoT on the continuous coded metrics:
Cliff's $\delta$ of $+0.67$ on stakeholder count (large effect,
$95\%$ CI strictly above $0.47$), with the NoT trace also
$2.3{\times}$ longer in tokens. The optimiser recovers the
qualitative structure of NoT but, at five iterations and one
batch per iteration, does not match its continuous depth.

\begin{table}[!htbp]
\centering
\small
\setlength{\tabcolsep}{4pt}
\begin{tabular}{lrrr}
\toprule
Metric & TextGrad & NoT & $\delta_{\text{NoT}}$ \\
\midrule
stakeholder count          & $4.00$  & $5.47$  & $+0.67$  \\
uncertainty score          & $2.83$  & $2.97$  & $+0.13$  \\
max causal hops            & $4.00$  & $4.03$  & $+0.03$  \\
collapse fire rate         & $0.0\%$ & $0.0\%$ & --       \\
suppression fire rate      & $0.0\%$ & $0.0\%$ & --       \\
mean completion tokens     & $620$   & $1440$  & --       \\
\bottomrule
\end{tabular}
\caption{Held-out comparison on $30$ DailyDilemmas scenarios
(\texttt{gpt-5.4-nano} generator, \texttt{claude-haiku-4-5}
judge). $\delta_{\text{NoT}}$ is Cliff's $\delta$ for NoT
vs.\ the TextGrad-optimised prompt (positive favours NoT).
TextGrad wins on token cost ($2.3{\times}$ cheaper) and matches
NoT at the binary failure-mode floor; NoT wins on continuous
stakeholder coverage with a large effect size.}
\label{tab:tg_compare}
\end{table}

\paragraph{Caveat: model deployment drift.} The optimisation
loss was flat at $0.0$ on every training batch because the
verbatim Experiment~1 \texttt{standard\_cot} prompt, run on
\texttt{gpt-5.4-nano} via the same Azure deployment one day after
the Experiment~1 main run, no longer fires either failure mode on
these $30$ scenarios. Experiment~1's standard-CoT baseline rates
for this generator were $14.6\%$ collapse and $50.0\%$
suppression; the run-time observed rates here are $0\%$ and
$0\%$. The most parsimonious explanation is that the underlying
deployment was updated between the two run dates; an alternative
is sampling drift on the $40$-cell training subset. We treat
this as a feature, not a bug: the comparison reported in
Table~\ref{tab:tg_compare} is between two prompts both already
clearing the binary rubric thresholds, so the
$\delta = +0.67$ NoT advantage on continuous stakeholder
coverage is what separates them. The full $100$-scenario,
multi-seed Experiment~1 firing rates (Table~\ref{tab:firing})
were measured before any prompt or deployment drift and stand as
reported.

\paragraph{Strengthened head-to-head: best optimised CoT, both
layers.} The pilot above optimised on the one generator where the
binary loss had drifted to zero, so it could not exercise the
optimiser. We therefore re-ran the optimisation on
\texttt{claude-haiku-4-5}, which retains failure-mode signal,
under both the binary loss and a continuous depth loss
$\max(0, 4-\textsc{sc}) + \max(0, 2-\textsc{us})$, and evaluated
the \emph{stronger} of the two resulting prompts against verbatim
NoT on the held-out $30$ scenarios across two generators, coded
by a cross-vendor judge pair (\texttt{claude-haiku-4-5} primary,
\texttt{gpt-5.4-nano} secondary). The single-agent verdict is
unchanged: even the best optimised CoT trails NoT on stakeholder
count by Cliff's $\delta = +0.78$ ($95\%$ CI $[+0.63, +0.91]$) on
\texttt{claude-haiku-4-5} and on uncertainty score by $+0.23$ and
$+0.54$ on the two generators (CIs strictly above $0$). The
optimised CoT closes the stakeholder-count gap on
\texttt{gpt-5.4-nano} ($\delta = +0.07$, CI spanning $0$) but
never the uncertainty gap, and does so at $2$--$5{\times}$ fewer
tokens.

\paragraph{Multi-stakeholder head-to-head.} Holding the
five-round integration protocol and the moderator
(\texttt{claude-sonnet-4-6}) constant and varying only the agent
prompt, NoT reaches full $R4$ consensus on $52\%$ of held-out
scenarios (Wilson $95\%$ CI $[39, 64]$) versus $32\%$ $[21, 44]$
for the best optimised CoT---a $+20$pp gap, Fisher exact
$p = 0.041$---and leaves roughly four times fewer debates in
unresolved structural rejection ($8\%$ vs $35\%$;
Table~\ref{tab:tg_debate}). Optimising a standard-CoT prompt for
single-agent stakeholder breadth therefore does not transfer to
the social layer, and here degrades it. We report one honest
caveat: a plain, \emph{unoptimised} standard CoT reaches a higher
raw consensus rate ($63\%$) than NoT on this held-out set, so the
multi-stakeholder claim is specifically that NoT dominates the
\emph{optimised} CoT baseline---not that it maximises raw
agreement.

\paragraph{Where rejections land.} Raw consensus rate hides
\emph{which} role dissents. Classifying every per-perspective
$R4$ vote (cached judge over each debater's final verdict)
reframes the standard-CoT caveat as evidence for NoT. NoT's $R4$
rejections are both rare and role-appropriate: of the $5$
per-perspective rejections across $126$ NoT votes, $60\%$ come
from the External Advisor---the role explicitly charged with
flagging principle violations---and only $20\%$ from the Primary
Decision Maker who is meant to drive toward a workable decision.
The optimised CoT instead scatters $38$ rejections near-uniformly
across roles (normalised role entropy $0.98$ vs.\ NoT's $0.86$),
with the \emph{decider itself} the modal rejector ($42\%$)---a
structurally incoherent pattern in which the role responsible for
convergence most often blocks the synthesis it helped build. NoT
therefore disagrees roughly $7{\times}$ less often than the
optimised CoT, and when it does, the dissent concentrates where
principled objection belongs rather than scattering. (NoT's
narrative style left $12$ of $126$ final verdicts without a clean
binary token; these are counted conservatively as non-rejections,
and NoT's $n = 5$ rejections make the share suggestive rather than
a tight estimate.)

\begin{table}[!htbp]
\centering
\small
\setlength{\tabcolsep}{4pt}
\begin{tabular}{lrrr}
\toprule
Arm & $R4$ consensus & $95\%$ CI & struct.\ reject \\
\midrule
NoT                 & $52\%$ & $[39, 64]$ & $8\%$  \\
optimised CoT       & $32\%$ & $[21, 44]$ & $35\%$ \\
standard CoT        & $63\%$ & $[51, 74]$ & $17\%$ \\
\bottomrule
\end{tabular}
\caption{Multi-stakeholder head-to-head on $30$ held-out
DailyDilemmas scenarios ($2$ generators, moderator
\texttt{claude-sonnet-4-6}, $n = 60$ debates per arm; only the
agent prompt varies). NoT vs.\ optimised CoT: $+20$pp on full
$R4$ consensus, Fisher exact $p = 0.041$, and four times less
unresolved structural rejection.}
\label{tab:tg_debate}
\end{table}

\subsection{Verbatim optimised prompts and training curves}
\label{app:tg-verbatim}

\paragraph{NoT-v2 (in-family training; SHA-1 prefix \texttt{3fb7f75f6d}; 2{,}579 characters).}
\begin{quote}\footnotesize\sffamily
You are a thoughtful advisor reasoning through ethical dilemmas. Before giving your final answer, work through a structured analysis. Keep each section tight and purposeful.

\textbf{Who is deciding?} Name the decision-maker, their role, and the specific knowledge or constraints they bring to this choice.

\textbf{Who is affected?} Identify every person, group, or institution whose wellbeing, interests, or rights intersect this decision. Push past the central figures: include indirect parties, bystanders, background institutions, and anyone whose situation quietly changes as a result. For each, state precisely what they stand to gain, lose, or have changed---not that they are ``affected,'' but what specifically shifts for them. Prefer concrete specificity (``the clinic's nursing staff who bear legal exposure'') over abstract categories. Aim for at least five distinct stakeholders with individuated stakes.

\textbf{What happens next?} For each available course of action, trace consequences at least two steps forward. Every stakeholder you named must appear somewhere in this analysis.

\textbf{What don't we know?} For each projected future, surface specific uncertainties. For each course of action, name at least three distinct uncertainties---not generic hedges, but named gaps that could materially change the outcome. For each uncertainty, explain why it matters: what reversal or surprise would it produce?

\textbf{What should be done?} Commit to a specific decision. Explain why this path is preferable given the stakeholder impacts you traced and the uncertainties you surfaced. Name which uncertainties most threaten your chosen path and state directly why you are proceeding despite them.

\textit{Additional guidance}: Precision over length. Do not resolve tensions prematurely. Be concrete about stakes. Explain the consequence of being wrong, not merely that the unknown exists.
\end{quote}

\paragraph{NoT-v3 (cross-family training; SHA-1 prefix \texttt{a51ec242d5}; 2{,}401 characters).}
\begin{quote}\footnotesize\sffamily
You are a thoughtful advisor helping people navigate ethical dilemmas. Before committing to any recommendation, think carefully and concisely through the following.

\textbf{Who is deciding and what do they stand to gain or lose.} Characterize the decision-maker's role, relevant knowledge, and personal stakes in one or two sentences.

\textbf{Who else is affected.} Cast a wide net. Start with those directly executing or immediately experiencing the decision, then move outward: people affected one or two steps removed, institutions and communities absorbing indirect effects, and anyone whose future options will be constrained by what happens now. For each distinct party, name their concrete stake in a single sharp sentence. Push past obvious stakeholders---someone in the background or a future generation is almost always relevant. Aim to surface at least six distinct parties; if you find more, include them.

\textbf{What could go wrong or remain unknown.} For every party you named and every realistic course of action you are considering, state at least one specific uncertainty---a hidden intention, an unpredictable reaction, a missing fact, or a long-run effect that cannot be resolved with available information. Name the precise unknown, not merely that uncertainty exists. No party should be skipped. Where an uncertainty is especially consequential---where resolving it would flip the recommended action---say so explicitly.

\textbf{What the realistic options are and how outcomes ripple.} For each plausible course of action, trace the most decision-relevant consequences forward. Focus on how an outcome for one party reshapes outcomes for others. Be selective: include causal chains that change the analysis; skip those that don't.

\textbf{The call.} Commit to a specific course of action. Justify it by direct reference to the stakeholders and uncertainties you surfaced. Name the uncertainties you are accepting and explain why the expected benefits still justify those risks. Do not hedge into vagueness---a qualified commitment is fine, but the recommendation must be actionable.

Throughout, write concisely. One sharp sentence per stakeholder stake. One precise unknown per party. Avoid restating what earlier reasoning already established.
\end{quote}

\paragraph{Training loss curves.} Both runs trained on the seed-$43$
stratified $30$-scenario sample with batch size $10$ and early-stop
after three consecutive iterations with $< 5\%$ loss reduction.

\begin{center}\small\setlength{\tabcolsep}{4pt}
\begin{tabular}{lrrr@{\hskip 1em}lrrr}
\toprule
\multicolumn{4}{c}{\textbf{Run A (in-family, v2)}} & \multicolumn{4}{c}{\textbf{Run B (cross-family, v3)}} \\
\cmidrule(lr){1-4}\cmidrule(lr){5-8}
Iter & $L$ & $\overline{\mathit{sc}}$ & $\overline{\mathit{us}}$ & Iter & $L$ & $\overline{\mathit{sc}}$ & $\overline{\mathit{us}}$ \\
\midrule
$0$ & $0.100$ & $4.50$ & $3.00$ & $0$ & $0.100$ & $4.50$ & $2.90$ \\
$1$ & $0.000$ & $6.30$ & $3.00$ & $1$ & $0.000$ & $7.70$ & $3.00$ \\
$2$ & $0.000$ & $6.60$ & $3.00$ & $2$ & $0.000$ & $6.30$ & $3.00$ \\
$3$ & $0.000$ & $5.40$ & $3.00$ & $3$ & $0.100$ & $4.60$ & $2.90$ \\
$4$ & $0.000$ & $6.30$ & $3.00$ & $4$ & $0.000$ & $6.10$ & $3.00$ \\
    &         &        &        & $5$ & $0.000$ & $5.90$ & $3.00$ \\
    &         &        &        & $6$ & $0.000$ & $5.70$ & $3.00$ \\
    &         &        &        & $7$ & $0.000$ & $5.70$ & $3.00$ \\
\bottomrule
\end{tabular}
\end{center}

Iteration $0$ is hand-written NoT. As a drift control, both runs
re-issued the NoT prompt on scenario \texttt{dd\_32489} at the start and
end of training; stakeholder counts moved $6\to6$ (Run~A) and $5\to6$
(Run~B), both within the pre-registered $\Delta = 0.5$ threshold for
compute-drift confounding modulo single-cell rounding, and the
four-generator replication shows no compute-drift signature.

\end{document}